\documentclass[sigconf]{acmart}
\usepackage{tabularx}
\usepackage{lscape}  
\usepackage{graphicx}
\usepackage{amsmath}
\usepackage{makecell}

\usepackage{amssymb}
\usepackage{ragged2e}

\usepackage{booktabs}
\usepackage{multirow}
\usepackage{array}
\usepackage{lipsum}  
\usepackage{booktabs} 
\usepackage{enumitem}
\usepackage{ragged2e}
\usepackage{float}
\usepackage{placeins}
\usepackage{graphicx}
\usepackage{subcaption}
\usepackage{array}

\newcolumntype{P}[1]{>{\raggedright\arraybackslash}p{#1}}

\AtBeginDocument{%
  }

\copyrightyear{2026}
\acmYear{2026}
\setcopyright{cc}
\setcctype{by}
\acmConference[CHI '26]{Proceedings of the 2026 CHI Conference on Human Factors in Computing Systems}{April 13--17, 2026}{Barcelona, Spain}
\acmBooktitle{Proceedings of the 2026 CHI Conference on Human Factors in Computing Systems (CHI '26), April 13--17, 2026, Barcelona, Spain}
\acmPrice{}
\acmDOI{10.1145/3772318.3791197}
\acmISBN{979-8-4007-2278-3/2026/04}

\begin{document}

\title[Signals of Success and Struggle]{Signals of Success and Struggle: Early Prediction and Physiological Signatures of Human Performance across Task Complexity}


\settopmatter{authorsperrow=2}

\author{Yufei Cao}
\orcid{0000-0002-9616-9913}
\affiliation{%
  \department{School of Computing}
  \institution{The Australian National University}
  \city{Canberra}
  \state{ACT}
  \country{Australia}}
\email{yufei.cao1@anu.edu.au}

\author{Penny Sweetser}
\orcid{0000-0002-6543-557X}
\affiliation{%
  \department{School of Computing}
  \institution{The Australian National University}
  \city{Canberra}
  \state{ACT}
  \country{Australia}}
\email{penny.kyburz@anu.edu.au}

\author{Ziyu Chen}
\orcid{0009-0009-0947-3887}
\affiliation{%
  \department{Computational Media Lab}
  \institution{The Australian National University}
  \city{Canberra}
  \state{ACT}
  \country{Australia}}
\email{ziyu.chen@anu.edu.au}

\author{Xuanying Zhu}
\orcid{0000-0002-3463-1447}
\affiliation{%
  \institution{The Australian National University}
  \city{Canberra}
  \state{ACT}
  \country{Australia}}
\email{xuanying.zhu@anu.edu.au}

\renewcommand{\shortauthors}{Cao et al.}

\begin{teaserfigure}
  \centering
  \includegraphics[width=\linewidth]{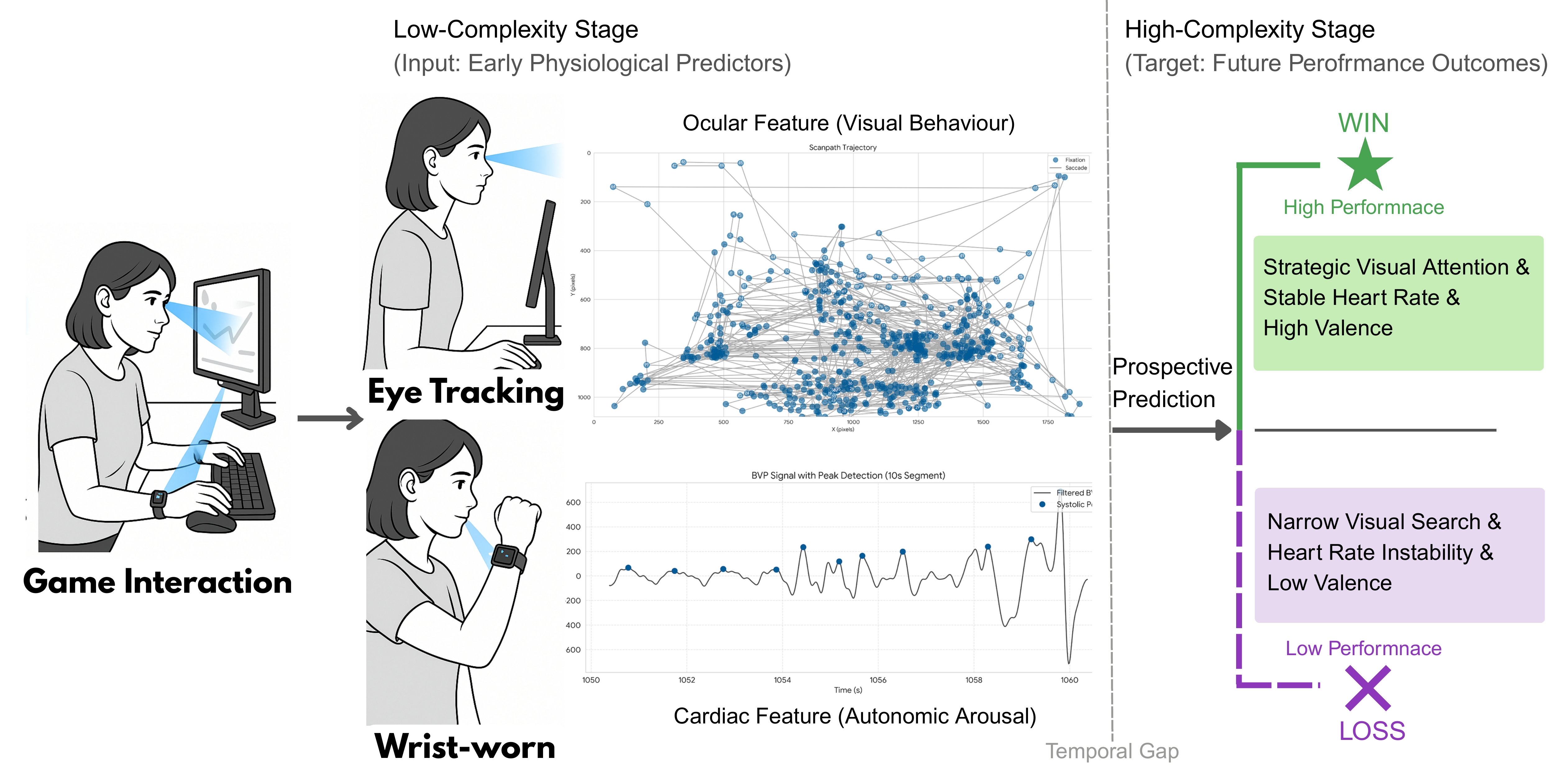}
  \vspace{-20pt}
  \caption{Overview of the prospective prediction pipeline. Early ocular and cardiac features from the low-complexity game session are used to predict performance outcomes in the subsequent high-complexity session. Ocular features capture visual behaviour, while cardiac features reflect autonomic arousal. Differences in visual patterns, autonomic profiles, and affective states are observed between high- and low-performing groups.
  }
  \Description[Overview of the prospective prediction pipeline.]{The figure shows how early physiological signals predict later game performance. A participant plays a PC game while eye tracking and a wrist-worn sensor record visual behaviour and cardiac activity. Example plots in the middle illustrate ocular and cardiac features from the low-complexity session. These early features feed into a prospective prediction step that forecasts outcomes in the later high-complexity session. The right side shows two possible outcomes: high performance, linked to strategic visual attention and stable heart rate, and low performance, linked to narrow visual search and heart-rate instability.}
  \label{fig:main_findings}
\end{teaserfigure}

\begin{abstract}
User performance is crucial in interactive systems, capturing how effectively users engage with task execution. Prospectively predicting performance enables the timely identification of users struggling with task demands. While ocular and cardiac signals are widely used to characterise performance-relevant visual behaviour and physiological activation, their potential for early prediction and for revealing the physiological mechanisms underlying performance differences remains underexplored. We conducted a within-subject experiment in a game environment with naturally unfolding complexity, using early ocular and cardiac signals to predict later performance and to examine physiological and self-reported group differences. Results show that the ocular–cardiac fusion model achieves a balanced accuracy of 0.86, and the ocular-only model shows comparable predictive power. High performers exhibited targeted gaze and adjusted visual sampling, and sustained more stable cardiac activation as demands intensified, with a more positive affective experience. These findings demonstrate the feasibility of cross-session prediction from early physiology, providing interpretable insights into performance variation and facilitating future proactive intervention.
\end{abstract}


\maketitle
 
\section{Introduction}

User performance is a key indicator of interaction effectiveness, capturing how successfully and efficiently users achieve task goals \citep{ISO9241-11, shneiderman1979human, egan1988individual, sonderegger2010influence}. This metric is critical across interactive contexts, ranging from high-risk operations such as aviation and driving, where reliable performance is paramount for operational safety \citep{binias2020prediction, zhou2020driver, Sarlija2020TaskPrediction, luzzani2024review}, to engagement-driven environments including educational platforms and digital games, where sustained achievement supports user motivation and ongoing participation \citep{mekler2013measuring, nacke2017gamification, mekler2017towards}. Beyond functioning as terminal outcomes, performance measures reflect the quality of coupling between user and system during task execution, signalling users' effective coping or emerging struggle in response to task complexity.
Anticipating performance outcomes enables interactive systems to provide prospective assessment and support intervention before performance breakdown occurs. This capability is essential for maintaining task continuity as task complexity escalates. Consequently, there is a growing need for interactive technologies to move beyond reacting to manifest errors towards predicting future performance trajectories \citep{akccapinar2019using, guo2025pptp, wei2025predicting}.

Prospective performance forecasting has proven feasible in real-world systems, primarily relying on behavioural metrics (e.g., data logs, user motion, task completion time) \citep{smerdov2023ai, akccapinar2019using, azizah2024predicting, guo2025pptp}. For instance, early predictive models in education can identify students at risk of failing based on their first weeks’ learning activities \citep{akccapinar2019using}, and game telemetry in esports has been used to predict the onset of acute performance decline during competitive play \citep{wei2025predicting}. However, while demonstrating the feasibility of early forecasting of performance trajectories, these behavioural indicators primarily capture what actions are taken but remain opaque with regard to the continuous cognitive processes and hidden physiological costs deployed to sustain performance, such as the mental effort or stress reactivity occurring between observable actions \citep{Doherty2018EngagementHCI, Winne2000SelfRegulatedLearning, Siemens2012LearningAnalytics, Li2022HandwrittenNotesLAK, Siemens2012LearningAnalytics}. This explanatory limitation motivates the use of measures that reflect the covert cognitive and physiological processes supporting task performance.

Ocular and cardiac signals fulfil these requirements by capturing the dynamics of visual behaviour and autonomic activity. Specifically, ocular measures map the spatiotemporal features that reveal how users explore, prioritise, and adapt to task-relevant elements, thereby offering a window into the internal cognitive processes guiding these ongoing actions \citep{konig2016eye}. Prior work has successfully leveraged eye-movement behaviour to study efficient visual search strategies and attentional guidance in complex interfaces \citep{video_game_scenery_analysis, tatler2017eye, hutton2008cognitive}, identify visual scanning patterns characteristic of experts versus novices \citep{ooms2014study, fogarty2023eye, dogusoy2014cognitive}, and quantify the mental effort required to maintain performance via pupil dynamics \citep{van2018pupil, zagermann2016measuring, gavas2017estimation, peysakhovich2015pupil}. Complementarily, cardiac measures (e.g., heart rate variability) index autonomic regulation, reflecting how users mobilise physiological resources to manage arousal and stress reactivity under fluctuating task demands \citep{jervcic2020modeling, luque2016heart, Mehler2012Sensitivity, zhou2017indexing}. Ocular and cardiac signals jointly capture the continuous interplay between visual information sampling and autonomic regulation that is relevant for task performance, establishing them as promising candidates for early prediction of subsequent performance outcomes. However, it remains unclear whether they are capable of prospective prediction under escalating task complexity in real interaction settings.

Despite the widespread use of ocular and cardiac measures to characterise expertise, workload, and self-regulatory capacity across a range of tasks \citep{ooms2014study, fogarty2023eye, laborde2014ability, hansen2003vagal, luque2016heart}, these applications have primarily focused on identifying users' momentary physiological states to interpret performance variations within the current task conditions. 
However, such momentary states are highly sensitive to local events and do not reliably indicate the user's overall level of task performance \citep{10.1145/3472307.3484674, faber2024effects}, particularly in real-world interactive systems where task demands naturally intensify and interaction proceeds
continuously rather than in brief, isolated intervals.
To address the gap in research where the early prediction of performance outcomes using physiological signals under evolving task complexity remains underexplored, we propose to prospectively predict performance outcomes using early ocular and cardiac signals under increasing task complexity. We situate our investigation in a digital game environment, which offers an ecologically valid yet controlled setting where performance outcomes (win/loss) arise naturally. These outcomes emerge as complexity progressively escalates, allowing performance to be examined across different levels of task demand. The engaging nature of games sustains participant involvement across continuous sessions and yields authentic data. Game performance outcomes also entail core processes of planning, problem solving, and strategy adaptation that generalise beyond entertainment, making games a powerful testbed for predicting performance outcomes. In this context, we operationalised escalating demands as two complexity stages: a low-complexity stage followed by a high-complexity stage, providing a structured basis for examining how early signals relate to later outcomes.

To complement these physiological indices, subjective affective ratings of valence, arousal, and dominance provide additional context on participants' affective experience and sense of control during gameplay. These measures help situate the observed physiological patterns within the self-reported affective states across the temporal progression of gameplay. By integrating these measures, our study asks:
\begin{itemize}
    \item \textbf{RQ1.} To what extent can early ocular and cardiac signals predict later performance outcomes as task complexity escalates?
    \item \textbf{RQ2.} How do ocular and cardiac features differ between high- and low-performing groups under increasing task complexity?
    \item \textbf{RQ3.} How do subjective affective experiences and sense of control differ between high- and low-performing groups?
\end{itemize}

To address these research questions, we conducted a within-subject experiment with 35 participants, recording ocular and cardiac signals while they played a deck-building game consisting of two sessions of increasing complexity. After each session, participants completed affective self-report scales. From the ocular data, we extracted saccade, pupil, and gaze-distribution features, and from the cardiac data, we derived blood volume pulse and heart rate indicators, following established links between these measures and task performance \citep{hayes2016pupil, liversedge2000saccadic, long2023heart, Holmqvist2011, zhou2017indexing}. To assess the predictive performance of ocular and cardiac signals from the low-complexity session in classifying performance outcomes in the subsequent high-complexity session (win/loss), we evaluated ocular-only, cardiac-only, and decision-level fused features under a leave-one-subject-out (LOSO) cross-validation scheme.

Our results show that the fusion model achieved a balanced accuracy of 0.86, demonstrating reliable early prediction of later performance. The ocular-only model performed comparably, whereas the cardiac-only model showed weaker predictive power (balanced accuracy = 0.70). Among the extracted features, gaze distribution, saccade dynamics, and heart rate were the most informative predictors distinguishing high- and low-performing participants. These measures exhibited different temporal trajectories across complexity sessions, providing interpretable context for performance-related differences in visual behaviour and autonomic activity. Self-reports further revealed significant differences in perceived emotion and sense of dominance, which supplemented the physiological evidence.

This study provides the first empirical evidence that physiological signals from the low-complexity session can forecast performance outcomes in the subsequent high-complexity session within a single game context, forming the basis for our contributions to multimodal psychophysiology computing, performance prediction, and applied machine learning in HCI by:
\begin{itemize}
    \item Demonstrating that early-session ocular and cardiac signals exhibit predictive value for later performance under escalating complexity within the examined game setting, extending prior physiological performance prediction research beyond concurrent state estimation.
    
    \item Identifying systematic differences between high- and low-performing groups in ocular indices of attentional control and cardiac indices of autonomic regulation across complexity sessions, characterising task-specific physiological patterns associated with performance regulation.
  
    \item Revealing subjective differences in affective experience and sense of control between high- and low-performing groups, enriching the interpretation of the physiological patterns across the progression of game complexity.
\end{itemize}

\section{Related Work}
We investigated the existing body of research on performance prediction in various interactive domains, on the use of multimodal physiological signals to capture performance-related user states, and on modelling approaches for physiological data. Prior studies show that behavioural data and physiological signals can forecast outcomes, but these approaches often rely on short temporal windows and give limited insight into the causes of performance differences. Psychophysiological work demonstrates that cardiac measures index internal regulation, pupil size reflects cognitive and affective states, and eye movements reveal attentional control, yet these signals are rarely combined. Modelling approaches for physiology-based prediction tasks benefit from tree-based ensembles and recurrent sequence models. This body of work shows how early physiological signals can be leveraged to anticipate performance outcomes in interactive tasks.

\subsection{Performance Prediction in Interactive Contexts}

Performance has been described through measurable outcomes of task execution, encompassing indicators of effectiveness, efficiency, and success under given conditions \citep{ISO9241-11, Endsley1997, Campbell1990}. As these outcomes reflect the quality of user–system interaction under task demands, recent research across interaction domains has increasingly investigated how predictive models can anticipate performance outcomes.

An increasing body of research in human–computer interaction has leveraged physiological sensing to predict task performance. \citet{heard2022predicting} mapped multimodal physiological signals to workload estimates, which in turn forecasted accuracy and efficiency in supervisory control tasks up to several minutes ahead. Similarly, \citet{guo2025pptp} combined ECG, GSR, EMG, and behavioural indicators (e.g., response timing) to anticipate trust levels in human–robot collaboration before task completion. In competitive eSports, prediction has been investigated at multiple temporal and physiological scales. \citet{smerdov2023ai} segmented eSports gameplay into temporal windows, using the combination of game logs, physiological signals, and chair-sensor signals to forecast player performance in subsequent segments, within 240 seconds ahead. \citet{minami2024prediction} instead focused on pre-match EEG activity, showing that gamma and alpha oscillations could forecast match outcomes, though interpretability remained constrained to neural correlates of preparation. Beyond predicting outcomes, \citet{wei2025predicting} operationalised the construct of “tilt” by detecting imminent breakdowns in eSports performance, demonstrating that sudden declines could be predicted seconds to minutes in advance from telemetry and physiological markers.

Across these domains, emerging evidence points to the value of early-task windows of physiological data for prospective prediction of performance outcomes. However, existing work has focused on state monitoring or short-term predictions (e.g., seconds ahead) within the same task conditions. While cardiac and neural signals are widely used, ocular measures, closely aligned with information processing and interface interaction, have been comparatively underexplored as predictors. This creates a need for approaches that (1) assess whether physiological predictors remain informative across phases where task complexity escalates, and (2) integrate ocular and cardiac measures to broaden the predictive scope of physiological sensing.
 
\subsection{Ocular and Cardiac Signals as Predictors of Performance}
Psychophysiological research is defined as the use of physiological signals to investigate psychological processes \citep{cacioppo2007handbook}. Among physiological channels, eye movements governed by the central nervous system provide rich indices of visual behaviour during information processing \citep{rayner1978eye}. Heart rate (HR), heart rate variability (HRV), blood volume pulse (BVP), and pupil diameter (PD) are regulated by autonomic nervous system (ANS) activity, indexing physiological arousal within cognitive and emotional contexts.

Eye movements in experimental settings typically comprise fixations and saccades, which represent the core oculomotor mechanisms enabling visual perception through periods of information uptake and rapid gaze shifts across the scene \citep{Holmqvist2011, konig2016eye}. They have been employed to study the direct reflection of external attention \citep{Stemmann2010, mack2003inattentional, engbert2015spatial, libby1973pupillary, bosse2007developmental}, visual search strategies \citep{salous2018investigating}, and interface interaction behaviours \citep{fatehi2022guiding, video_game_scenery_analysis}. The influential “eye–mind assumption” posits that what people fixate on reflects where cognitive processing occurs \citep{just1976}. Researchers have identified fixation and saccade features as reliable indicators of cognitive performance \citep{poole2006, bavcic2022advancing, Holmqvist2011}. Fixation duration and frequency have been identified as reliable indicators of cognitive load \citep{velichkovsky2019visual, pradhan2022cognitive} and attentional allocation \citep{tatler2017eye, negi2020fixation}. Saccadic behaviour further indexes visual information processing and processing efficiency \citep{liversedge2000saccadic, hutton2008cognitive, seassau2014binocular, findlay1997saccade}. Target selection during saccades is strongly shaped by high-level cognitive factors such as expected rewards and task relevance \citep{ooms2014study}, showing sensitivity to top-down modulation \citep{wolf2021vision}. Under high-load conditions, such as driving \citep{cardona2014blinking}, video gameplay \citep{mallick2016use}, or online interviews \citep{behroozi2018dazed}, saccades of larger amplitude and higher velocity have been associated with higher stress, increased task complexity, and reduced concentration. As cognitive demands intensify, saccadic efficiency deteriorates, leading to delayed initiation and increased error rates during visual search tasks \citep{liversedge2000saccadic}.

At the spatial level, AOI-based gaze distribution analysis has been widely used to examine how visual sampling is allocated across interfaces in interactive environments. Prior studies show that skilled users allocate a greater proportion of gaze to functionally critical regions. For example, in surgical simulation, experts focus more on anatomically important areas \citep{fogarty2023eye, li2023using}, and in map-reading tasks, skilled users concentrate fixations on semantically informative regions rather than distributing gaze diffusely across the display \citep{ooms2014study}. In game-based learning environments, gaze data have been used to track attention to instructional cues and gameplay elements, showing how fixation patterns characterise task-driven allocation of visual resources and support interface design decisions \citep{kiili2014eye}.

Cardiac measures provide a complementary index of autonomic activity during task execution. In cognitively demanding tasks, elevated heart rate has been observed alongside higher performance levels, though this benefit diminishes under the highest difficulty levels \citep{alshanskaia2024heart}. In serial-subtraction tasks, HR rises as participants maintain performance under greater cognitive demands \citep{kennedy2000glucose}. Similar associations have been reported in interactive contexts such as serious games, where HR increases with cognitive load in parallel with pupil dilation \citep{jervcic2020modeling}. However, in entertainment gameplay, HR also varies with experiential quality, in which higher HR is linked to greater tension and negatively valenced player experience \citep{drachen2010correlation}. HRV, defined as the fluctuation in time between successive heartbeats \citep{taskforce1996hrv}, has long been recognised as a sensitive index of ANS function \citep{hon1965electronic, kristal1995heart}. Greater resting HRV predicts better cognitive flexibility in older adults \citep{park2014resting} and supports working memory performance \citep{hansen2003vagal}. Extending these associations to interactive contexts, \citet{lan2018player} showed that short-term HRV indices could effectively predict player performance metrics, such as the number of kills, demonstrating the potential of HRV as a predictor of performance in interactive environments. In parallel, BVP, which reflects the volume of blood flowing through peripheral vessels, has been widely employed in studies of stress and cognitive load \citep{healey2005detecting}. \citet{zhou2017indexing} demonstrated that BVP values varied systematically with levels of cognitive load, showing a connection between BVP dynamics and mental demands.

Pupil diameter (PD) refers to the size of the pupil and has been consistently linked to autonomic activity. Classic work by \citet{hess1960pupil} first showed that pupils dilate in response to interesting relevant stimuli. \citet{libby1973pupillary} found that pupil dilation and cardiac deceleration co-occur when a stimulus elicits greater attention and interest, identifying PD as a sensitive indicator of visual orienting during stimulus processing. In addition, a substantial body of work has linked task-evoked PD to mental effort. Studies of cognitive control report that pupil size increases as mental demands rise, such as in adaptive learning \citep{sibley2011pupil, sayali2023learning} and cognitive-load manipulation \citep{gavas2017estimation}. These findings describe PD as a reliable marker of the effort involved in inhibiting responses and sustaining working memory \citep{van2018pupil}. Multimodal studies increasingly incorporate PD with other signals to improve the performance of classification models of user states \citep{zheng2014multimodal, 10.5555/2832249.2832411}.

Overall, these findings show that ocular measures of saccades, fixations, and gaze allocation capture how users visually access information and distribute attention during ongoing interaction, while cardiac measures provide insight into changes in autonomic activation across different task conditions. These modalities offer complementary perspectives on user state during interaction, supporting their use in models that predict task performance.
to characterise participants’ reported experiences.
\begin{figure*}[t]
    \centering
    \subfloat[Low-complexity session]{%
        \includegraphics[width=0.49\textwidth]{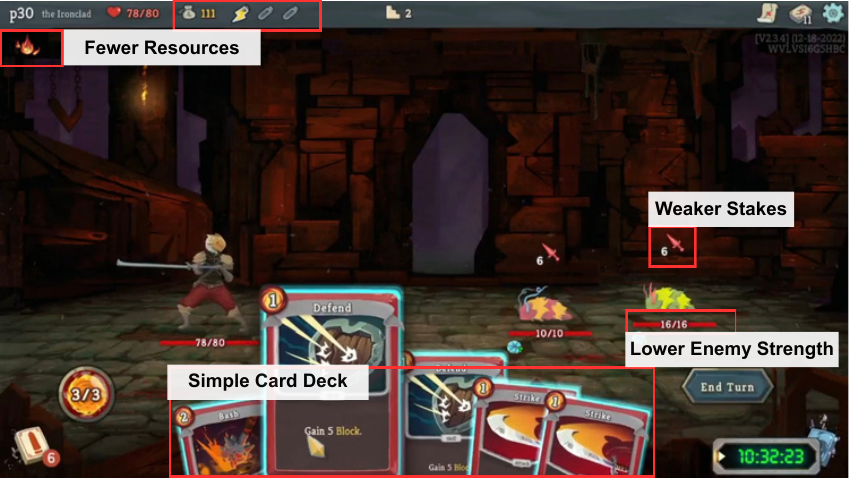}
    }
    \hfill
    \subfloat[High-complexity session]{%
        \includegraphics[width=0.49\textwidth]{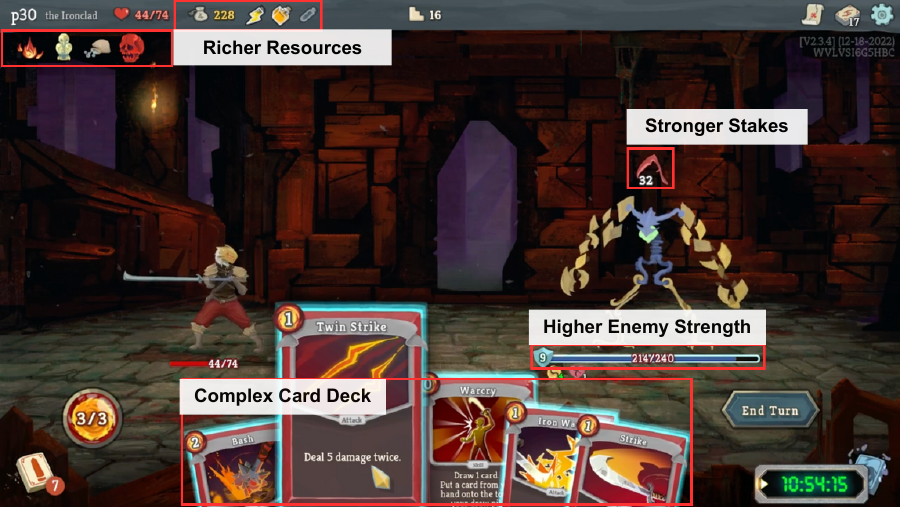}
    }

    \caption{Example game scenes illustrating the two experimental sessions. 
    (a) Low-complexity session, characterised by fewer resources, weaker enemies, and a simple card deck; 
    (b) High-complexity session, featuring more resources available, stronger enemies, and complex cards with intricate effects.}
    
    \Description[Screenshots of game scenes across low-complexity and high-complexity phases.]
    {The figure comprises two screenshots from the game, illustrating the core experimental phases. Panel (a) shows the low-complexity session, where the interface displays fewer items, an opponent with lower health, and simple card options. Panel (b) shows the high-complexity session, featuring a boss fight with more items visible, a stronger opponent, and cards with more complex effects.}
    \label{fig:game_scenes}
\end{figure*}
\section{Method}
We conducted a within-subject laboratory study in a digital game environment comprising two escalating complexity sessions. During the experiment, we collected ocular and cardiac signals for physiological modelling, as well as a post-session self-report of affective experience for complementary analysis. Using the ocular and cardiac data, we developed a prediction framework encompassing preprocessing, feature extraction, and model development to forecast performance outcomes. In addition, we examined the contribution of individual ocular and cardiac features and visualised their trends across complexity levels, and the subjective data were analysed separately 
\subsection{Participants}
We recruited 35 participants (16 male, 17 female, 2 non-binary) from the university campus and online platforms. Participants were aged 18–34 years, with 54.3\% in the 18–24 range and 45.7\% in the 25–34 range. The majority held a bachelor’s degree or higher. To ensure a comparable baseline, only individuals with no prior experience of the game were included. All participants reported normal or corrected-to-normal vision. The study protocol was approved by the university ethics committee, and all participants provided informed consent prior to participation. Participants received either course credit or a voucher as compensation.

\subsection{Experiment Design}
We conducted a within-subject laboratory experiment using a commercial video game, \textit{Slay the Spire} (Figure~\ref{fig:game_scenes}), dividing the gameplay into two sessions corresponding to lower and higher task complexity. 
A within-subject design was employed to determine whether early physiological signals could predict later performance under escalating complexity. By keeping the progression fixed for each participant and referencing physiological changes against their own low-demand baseline, the design isolates dynamic adjustments to rising complexity while minimising the influence of stable individual differences.

\textit{Slay the Spire} \citep{slaythespire} is a turn-based deck-building game built around a core loop of strategic card selection, resource management, and adaptive planning. Players act using a customisable deck of cards, where each card represents attacks, defences, or effect actions. During each combat encounter, enemy intent is visible, requiring players to choose cards in response to incoming threats, manage their limited energy and resources, and sequence actions efficiently. Across the run, players progress through successive encounters by building and refining their deck, culminating in a boss fight that concludes the level.

The game offers a progressive task structure in which combat demands escalate from normal battles to a final boss that appears at the end of the first level, creating a natural increase in overall task complexity as players advance (Figure~\ref{fig:game_scenes}, left: normal battle; right: boss battle). Early battles typically feature simpler decision requirements, fewer interacting resources, and enemies with straightforward attack patterns. In contrast, the boss encounter imposes substantially higher demands, featuring enemies with greater durability, more punishing attack cycles, and unique mechanics that disrupt the player's deck or introduce additional constraints, requiring more elaborate reasoning over a richer set of interdependent cards, resources, and combinational effects. 
Accordingly, the experiment distinguished between two sessions: a low-complexity session encompassing the pre-boss encounters, and a high-complexity session corresponding to the final boss encounter. This design reflects the game’s inherent progression rather than imposed manipulation of task difficulty, and creates a temporal structure in which early-stage physiological patterns under lower demands can be used to forecast performance outcomes under later, higher task complexity. 
This escalation aligns with cognitive load theory, in which task difficulty rises with increases in element interactivity \citep{sweller1994cognitive}. Early encounters involve low interactivity and offer a lower-load segment that precedes the sharp rise in cognitive demands, whereas later encounters require coordinating many interdependent elements, making performance increasingly sensitive to individual differences.

To ensure comparability across participants, we fixed the character selection and level map using the game's customisation options and applied a constant run seed, producing an identical layout and enemy sequence across all runs. The entire experiment used only the first level with a fixed map layout. Each participant began with a short built-in introductory sequence for familiarisation before entering the main combat stages. The low-complexity session lasted up to 25 minutes, whereas the high-complexity session lasted up to 15 minutes.
Each participant completed a single uninterrupted run. If the player’s health points were depleted at any point, the run terminated without restart and was coded as a loss. Reaching the final boss and defeating it was coded as a win. Participants were explicitly instructed that their goal was to defeat the boss, ensuring consistent motivation and engagement across the game progression.

\subsection{Apparatus and Signals}
\begin{figure}
    \centering
    \includegraphics[width=\linewidth]{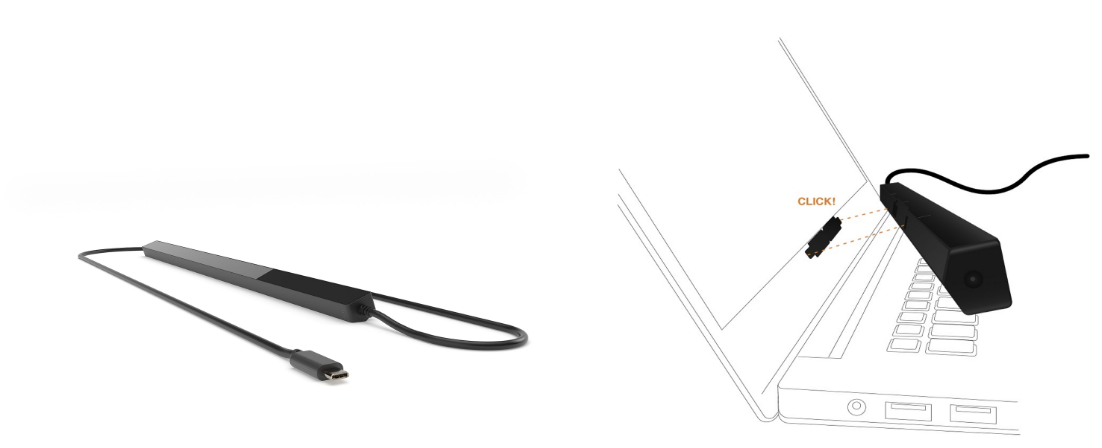}
    \caption{Eye tracking system (Tobii Pro Fusion).}
    \Description[Eye tracker used in the experiment.]
    {The figure shows the Tobii Pro Fusion eye tracker, a rectangular bar-shaped device designed to be positioned below a monitor to record eye movements.}
    \label{fig:eye}
\end{figure}

\begin{figure}
    \centering
    \includegraphics[width=0.45\linewidth]{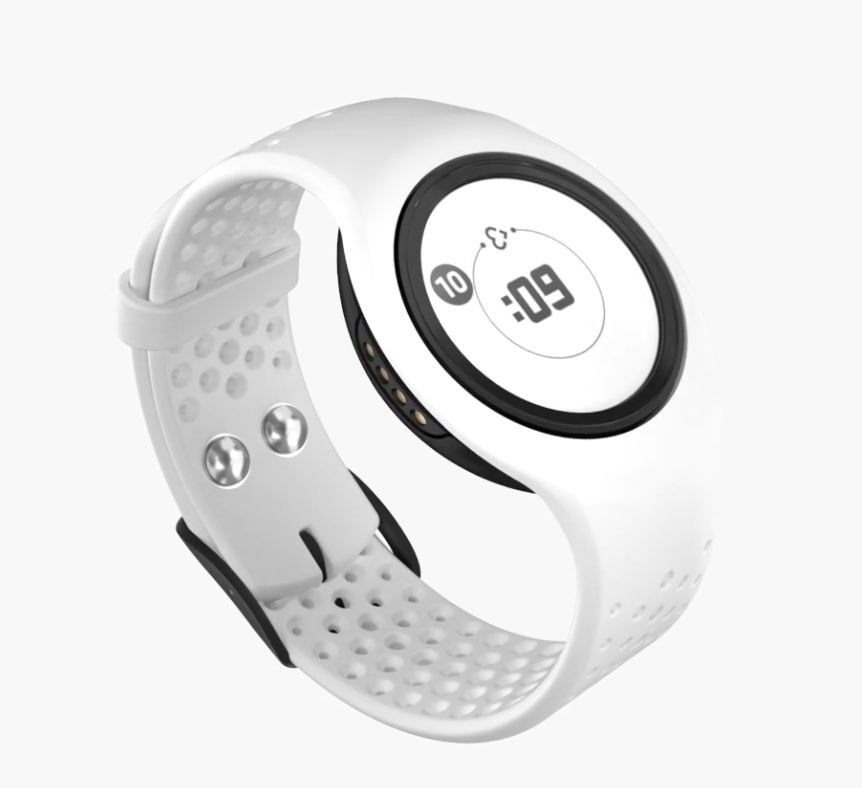}
    \caption{Cardiac recording device (Empatica EmbracePlus Wristband).}
    \Description[Cardiac recording device used in the experiment.]
    {The figure shows the Empatica EmbracePlus, a black wristband with sensors on the inner side, worn on the wrist to collect cardiac signals.}
    \label{fig:cariac}
\end{figure}

Ocular data were recorded with the Tobii Pro Fusion, a screen-based binocular eye tracker (see Figure ~\ref{fig:eye}) \citep{TobiiProFusion}. The device sampled at 250 Hz and offered high spatial precision. It was integrated with a 23.8-inch monitor (1920 × 1080 resolution), and participants were seated approximately 60 cm from the screen without a chin rest, allowing for natural head movement. We used Tobii Pro Lab software for calibration and data collection \citep{tobii_accuracy_precision}.
We selected a screen-based tracker to support natural interaction, ensuring stable gaze estimation while maintaining ecological validity for seated game play.

Cardiac data were recorded using the EmbracePlus, a wearable multisensor device developed by Empatica Inc (see Figure ~\ref{fig:cariac}) \citep{empatica_embraceplus}. 
The EmbracePlus records BVP via an optical photoplethysmography (PPG) sensor to capture cardiovascular pulse activity. BVP was sampled at 64 Hz, with data acquisition managed via the Empatica mobile app and uploaded to the Empatica Research Portal. 
We selected a wrist-worn device to ensure comfort and minimise intrusiveness, enabling long-duration recording without distracting participants during gameplay.
The wrist-based form factor also aligns with consumer smartwatches, and using cost-effective, readily accessible technology enables investigating whether physiological sensing commonly found in everyday smartwatches is capable of supporting performance prediction in interactive environments.

\subsection{Procedure}
\begin{figure*}[t]
    \centering
    \includegraphics[width=0.95\textwidth]{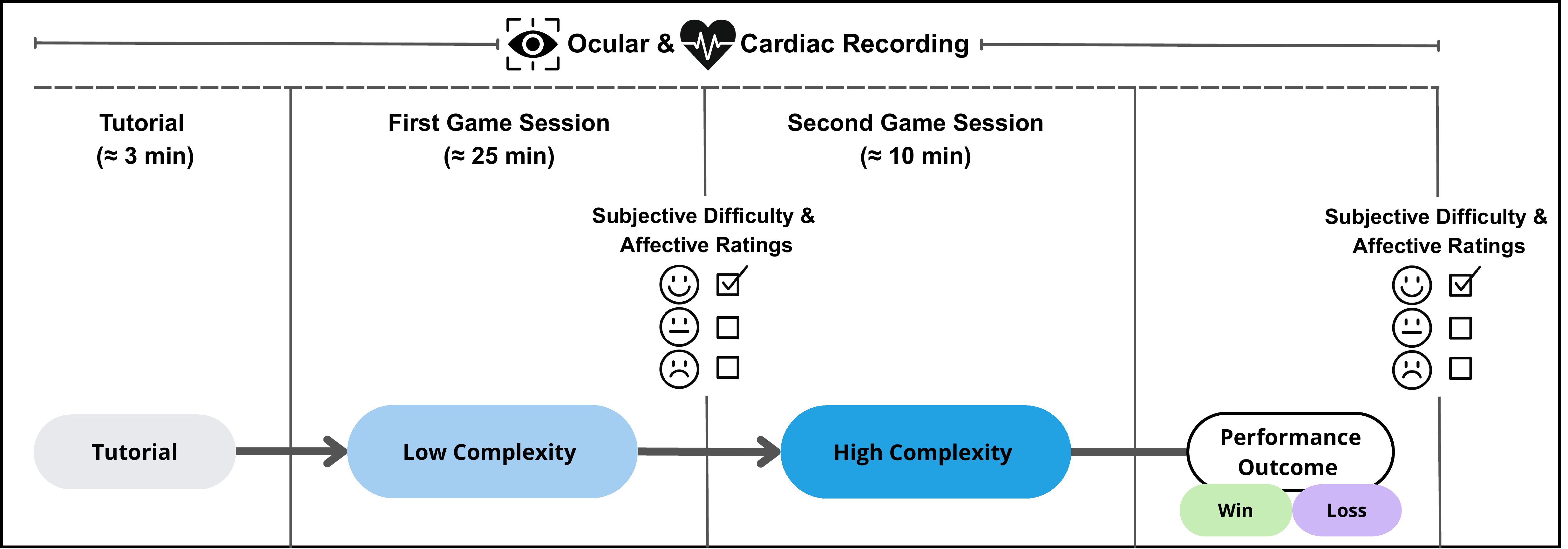}
    \caption{Experimental procedure with annotated session structure and physiological recording.}
    \label{fig:game_session_procedure}
    \Description[Experimental procedure showing sessions, physiological recordings, and self-reports.]
    {The figure presents the sequence of the study. Participants begin with a tutorial of about 3 minutes. They then enter the first game session, lasting about 25 minutes, which contains the low-complexity phase. A second game session of about 10 minutes follows and includes the high-complexity boss fight. Eye-tracking and cardiac signals are recorded continuously across both game sessions. After each session, participants provide subjective difficulty and affective ratings. The procedure ends with a performance outcome classified as win or loss.}
\end{figure*}

The overall structure of the experimental sessions is shown in Figure~\ref{fig:game_session_procedure}. Participants first completed a demographic questionnaire. At the beginning of the experiment, each participant was fitted with an Empatica wristband on the non-dominant wrist and then completed a 9-point calibration and validation procedure in Tobii Pro Lab. Calibration was repeated if the mean accuracy exceeded 1.0° or if data loss was greater than 10\%. All participants met these criteria before proceeding. The first game session began with a built-in tutorial, during which participants could ask questions about the game mechanics. Following this session, they completed a game difficulty scale and an affective experience scale. Before the second session, calibration was repeated to ensure eye-tracking quality. Finally, the second game session concluded with the same self-report questionnaires. 

\subsection{Subjective Measures}
Perceived game task difficulty was assessed using a 5-point Likert scale, where 1 indicated very easy and 5 indicated very hard. Participants rated the overall difficulty of both the low-difficulty and high-difficulty sessions. The questionnaire item was phrased as: "Please rate your overall experience regarding the difficulty level of the first game session." 

Affective experience was measured using the Self-Assessment Manikin (SAM) \citep{peacock1990stress}, a non-verbal pictorial instrument that assesses three dimensions of emotion. Valence captures the degree of pleasantness, ranging from unpleasant (1) to pleasant (5). Arousal reflects the level of physiological activation, ranging from calm (1) to excited (5). Dominance represents the perceived sense of control, ranging from low control (1) to high control (5). Participants were instructed to select the figure that best represented their emotional state during the session. The item was phrased as: “Please choose from the above figures which best matches your state when you play the game.”

\section{Signal Processing and Feature Engineering}
To enable predictive modelling from physiological input, we transformed raw ocular and cardiac signals into structured features via a multi-stage pipeline comprising denoising, segmentation, and modality-specific windowing, followed by feature extraction, fusion, and selection to prepare the data for classification.
\subsection{Feature Categories}
To structure the feature space for subsequent modelling, we categorised the features into ocular and cardiac groups, summarised in Table~\ref{tab:eye_features} and Table~\ref{tab:cardiac_features}. The full set of ocular and cardiac features, along with their statistical definitions, is provided in Appendix~\ref{app:feature}.

\begin{table}
\caption{Overview of extracted ocular features and their functional roles.}
\centering
\begin{tabularx}{\linewidth}{
l
>{\RaggedRight\arraybackslash}X
>{\RaggedRight\arraybackslash}X
}
\toprule
\textbf{Category} & \textbf{Feature Names} & \textbf{Functional Role} \\
\midrule
Fixation &
Fixation Duration, Fixation Count &
Periods where gaze remains relatively stable, indicating focused attention. \\
\midrule
Saccade &
Saccade Amplitude, Saccade Duration, Saccade Velocity, Saccade Count, Saccade to fixation ratio &
Rapid eye movements between fixations, reflecting visual scanning and search behaviour. \\
\midrule
Pupillometry &
Pupil Diameter &
Pupil size changes, associated with cognitive effort and emotional arousal. \\
\midrule
AOI Features &
AOI Hit Rate, AOI Proportion &
Distribution and allocation of visual attention across predefined regions of interest on the game interface. \\
\bottomrule
\end{tabularx}
\label{tab:eye_features}
\end{table}

\begin{table}
\caption{Overview of extracted cardiac features and their functional roles.}
\centering
\begin{tabularx}{\linewidth}{
>{\RaggedRight\arraybackslash}X
>{\RaggedRight\arraybackslash}X
>{\RaggedRight\arraybackslash}X
}
\toprule
\textbf{Category} & \textbf{Feature Names} & \textbf{Functional Role} \\
\midrule
Heart Rate Metrics &
Heart Rate, Heart Rate Range &
Reflect overall cardiovascular activation. \\
\midrule
Heart Rate Variability Metrics &
Root Mean Square of Successive Differences (RMSSD), Mean NN Interval &
Reflect autonomic nervous system regulation. \\
\bottomrule
\end{tabularx}
\label{tab:cardiac_features}
\end{table}
The ocular group consists of pupillometry, fixation, and saccadic behaviours, reflecting pupil-linked arousal and visual scanning dynamics \citep{van2018pupil, findlay1997saccade, negi2020fixation}. In addition, gaze distribution to predefined areas of interest (AOI) was included to capture attention allocation to task-critical interface regions that are essential for strategic choices \citep{Holmqvist2011}. AOIs were defined as several interface elements, including the hand cards (available actions), the top toolbars displaying potions (consumable items) and relics (persistent upgrades). A detailed illustration of these AOIs is provided in Appendix~\ref{app:aoi}.

The cardiac group consists of heart-related metrics, reflecting general arousal, autonomic regulation, and peripheral cardiovascular reactivity \citep{jorna1993heart, taskforce1996hrv, zhou2017indexing}. 


\subsection{Data Preprocessing}
Raw ocular and cardiac signals were preprocessed through a series of steps. First, all physiological time series were offline-aligned to the game timeline by synchronising the signal timestamps with a millisecond-precision on-screen clock plugin. Based on this alignment, each participant’s data were segmented into the tutorial, low-complexity, and high-complexity sessions. 

Raw pupil diameter data were cleaned to remove artifacts and sensor noise. Periods of signal loss (e.g., blinks), which accounted for less than 5\% of the dataset, were corrected using linear interpolation to maintain temporal continuity. A fourth-order Butterworth low-pass filter (cutoff = 4~Hz) was then applied to suppress high-frequency noise. For gaze metrics, raw screen-coordinate data (X, Y) were used to compute saccade kinematics. Velocity profiles were smoothed using a Savitzky--Golay filter \cite{SavitzkyGolay1964} with a 13-sample window and a third-order polynomial, reducing jitter and improving the robustness of velocity estimation. Data points located outside the display boundaries were removed to ensure that only valid on-screen gaze behaviour was analysed.

Raw BVP signals were filtered using a sixth-order Butterworth low-pass filter. A 3 Hz cutoff to remove high-frequency noise and motion-induced artifacts. Zero-phase forward–backward filtering (filtfilt) was applied to eliminate phase distortion, preserving the temporal alignment of systolic peaks with the original physiological events.

\subsection{Feature Extraction}
To transform the cleaned ocular and cardiac time series into interpretable measures, feature extraction was carried out separately for each modality.

\textbf{Ocular features} encompassed pupillometry measures, saccade, fixation metrics, and AOI-based gaze distribution. Pupil diameters were obtained directly in millimetres for both eyes. Saccade amplitude and velocity were calculated from the Euclidean distance between successive gaze points in the x–y plane relative to movement duration. Saccade and fixation counts were determined by marking the onset of each event, while fixation durations were derived from gaze duration values. AOIs were defined from in-game screenshots using automated image processing (OpenCV-based template matching and contour extraction) \citep{bradski2000opencv}, yielding pixel and normalised coordinates for two fixed toolbar elements. Gaze samples were compared against these AOI boundaries to produce two indices: the proportion of gaze samples within each AOI and the frequency of AOI hits. 
Based on preliminary testing of window lengths, we used a 500 ms sliding window with a 250 ms step for ocular feature calculations. This configuration yielded stronger predictive performance and provides a practical trade-off between measurement stability and the temporal resolution required to capture task-evoked changes \citep{peelle2021time}, while remaining sufficient for tracking rapid fluctuations in pupil size and eye-movement dynamics in our experimental settings \citep{Holmqvist2011}.

\textbf{Cardiac features} were derived from the BVP signals using the NeuroKit2 package 
\cite{neurokit, makowski2021neurokit2}. The package detects individual heartbeats and computes inter-beat intervals (IBIs), defined as the time between successive cardiac cycles \cite{pham2021heart}. From the IBIs, we calculated HR and two time-domain HRV indices: RMSSD and MeanNN. All cardiac features were computed in non-overlapping 15-second windows to ensure the inclusion of a sufficient number of cardiac cycles for estimating heart rate and time-domain variability metrics. Previous work has shown that 10-second ultra-short recordings provide acceptable accuracy for estimating these measures \cite{baek2015reliability, munoz2015validity}.

From these measurements, we extracted 35 features in total, comprising 29 ocular features and 6 cardiac features. Detailed computation procedures and statistical definitions for these ocular and cardiac features are provided in Appendix~\ref{app:feature}. 
 
Following the computation of ocular and cardiac features, a within-subject z-score normalisation was applied to standardise feature distributions, defined as subtracting each participant’s mean and dividing by their standard deviation across all windows.
This per-participant normalisation centred each feature distribution around zero and scaled it by its own variability, thereby removing baseline differences across individuals while preserving the relative fluctuations that reflect within-subject dynamics.
\begin{figure*}
    \centering
    \includegraphics[width=\textwidth]{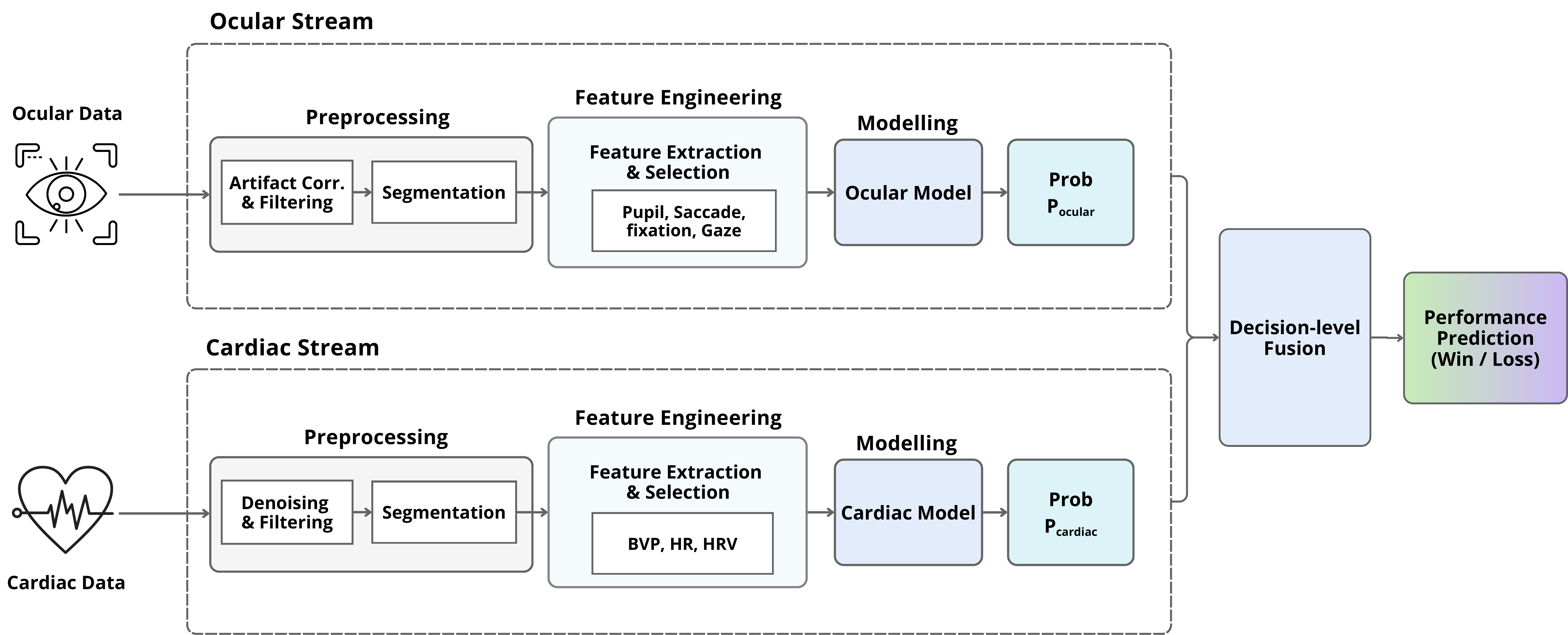}
    \caption{Processing framework integrating ocular, cardiac, and fused modalities.}
    \label{fig:framework}
    \Description[Framework from signal recording to model evaluation.]
    {The figure shows two parallel processing streams for ocular and cardiac data.
The top stream illustrates the ocular pathway: raw eye-tracking data undergoes artifact correction, filtering, and segmentation. Feature engineering then extracts pupil diameter, saccades, fixations, and gaze measures. These features are used to train an ocular model that outputs a probability of high or low performance.
The bottom stream illustrates the cardiac pathway: cardiac signals are denoised, filtered, and segmented. Then, feature engineering extracts blood volume pulse, heart rate, and heart-rate variability. These features train a cardiac model that outputs its own performance probability.
Both model outputs are combined through decision-level fusion to generate the final performance prediction of win or loss.}
\end{figure*}
\subsection{Feature Selection and Dataset Construction}
Feature selection was conducted iteratively using statistical filtering, model-based ranking, and ablation assessment.
We initiated with statistical filtering, where descriptive statistics and the Mann–Whitney U (MWU) test \cite{conover1999practical} were applied to the unimodal feature sets. Shapiro–Wilk tests \cite{peck2015introduction} and Quantile–Quantile (Q–Q) plots \cite{peck2015introduction} were used to assess normality prior to the MWU test, followed by tests for equal variance. In addition, we incorporated model-based feature importance scores from tree-based learners to prioritise informative predictors and remove redundant features.
Candidate subsets were then evaluated in ablation experiments to assess how the removal of specific feature categories affected generalisation performance. This iterative procedure guided the final retained features. The resulting multimodal feature set comprised 10 ocular features and 2 cardiac features. A complete summary of all extracted features and the final subset is provided in Appendix~\ref{app:feature_summary}. The final dataset contained 35 participant-level instances, with the performance classes nearly balanced (win: 19, loss: 16).

\section{Predictive Modelling}
To evaluate whether early physiological responses forecast subsequent performance, we built independent predictive models from ocular and cardiac features and then combined their outputs through a decision-level fusion framework. Generalisability was assessed using a LOSO protocol.
Figure~\ref{fig:framework} shows the processing pipeline for ocular, cardiac and fusion modalities. 
Both data streams undergo preprocessing, feature engineering, and model training independently. 
The ocular stream extracts pupil diameter, saccade, fixation, and gaze features, whereas the cardiac stream extracts features derived from BVP, HR, and HRV. 
Each modality produces a probability estimate ($P_{\text{ocular}}$, $P_{\text{cardiac}}$). 
The two probability outputs are then combined using a decision-level fusion strategy for final performance prediction (win or loss).

\subsection{Classification Models}
We trained three supervised classifiers: XGBoost \citep{chen2016xgboost}, CatBoost \citep{prokhorenkova2018catboost}, and Linear SVM \citep{cortes1995support}.
Tree-based ensembles (XGBoost, CatBoost) were selected for their ability to model non-linear interactions in physiological data. Additionally, a Linear SVM was trained as a baseline to assess whether the decision boundary could be linearly separated.
Hyperparameters for each model (e.g., tree depth, learning rate, regularization strength) were optimised using grid search within each training fold. The detailed tuning ranges and final configurations are provided in Appendix~\ref{app:hyperparams}.

\subsection{Decision-Level Late Fusion}
To combine the predictions from ocular and cardiac models, we adopted a decision-level late fusion. This approach is appropriate for multimodal signals that differ in temporal resolution, statistical properties, and windowing structures. Decision fusion aggregates independently trained model outputs and therefore accommodates heterogeneous physiological modalities without requiring shared representations \citep{kittler2002combining,lahat2015multimodal}.

Our fusion approach follows a consensus-first strategy \citep{alzubi2018consensus, du2025foundations, zhou2025ensemble}. A \emph{consensus} is identified when the two modalities independently produce the same subject-level label $(\hat{y}_{\text{ocular}} = \hat{y}_{\text{cardiac}})$; this prediction is retained as a high-confidence agreement. These consensus subjects retain their agreed-upon labels without further fusion. Only disagreement cases are passed to downstream fusion operators. This strategy concentrates fusion capacity on conflicts between modalities and reduces unnecessary decision noise, consistent with the broader ensemble-learning principle that classifier agreement implies reliability \citep{zhou2025ensemble}.

For each subject, both modalities output a probability $p$ and a subject-specific decision threshold $\tau^{\*}$ chosen on the training folds to maximise balanced accuracy. A hard decision was obtained by thresholding $\tau^{\*}$, i.e., $\hat{y} = 1$ if $p \ge \tau^{\*}$ and $0$ otherwise. The use of personalised thresholds $\tau^{\*}$ ensures that each individual’s decision boundary reflects their own score distribution.

Conflict subjects $(\hat{y}_{\text{ocular}} \neq \hat{y}_{\text{cardiac}})$ undergo fusion. For these cases, we employ a stacking-based late fusion approach. 
A meta-classifier (XGBoost) learns a non-linear mapping from the two modality-specific confidence scores $[p_{\text{ocular}}, p_{\text{cardiac}}]$ to a fused score estimate, following established methods of stacked generalisation \citep{wolpert1992stacked, zhou2025ensemble}.
A fusion-specific threshold $\tau^{*}_{\text{fusion}}$ is then re-estimated on the training folds to maximise balanced accuracy, calibrating the fused decision boundary to the score distributions of conflict cases. 
\subsection{Cross-Participant Evaluation (LOSO)}
We adopted a LOSO across-subject evaluation protocol to assess model generalisability.
In each fold, one participant was held out for testing, and the model was trained on data from all remaining participants. Inference was performed using only the held-out participant’s low-complexity session signals to predict their performance outcome in the subsequent high-complexity session, providing a strict test of prospective prediction under escalating task demands.
Within each training fold, hyperparameter tuning and decision-threshold optimisation were performed through an inner participant-wise cross-validation loop on the training set, selecting the threshold that maximised balanced accuracy.
After tuning, the model was retrained on the full training fold and evaluated on the held-out participant. Class imbalance between win and loss outcomes was addressed by applying class weighting within the tree-based models.
Finally, window-level probabilities were aggregated via logit-mean pooling to produce a single participant-level prediction. 
Performance was reported using Balanced Accuracy (BAcc), Macro-F1, Macro-Precision (Macro-Pr), Macro-Recall (Macro-Re), and Matthews Correlation Coefficient (MCC).

\section{Results} 
We first present the self-reported task complexity and affective experience measures. We then evaluate the predictive performance of ocular, cardiac, and fused models. Finally, we analyse the key ocular and cardiac predictors across performance groups, illustrating both overall differences and their temporal trajectories across task stages.

\subsection{Self-reported Task Difficulty and Affective Experience across Complexity Levels}
\begin{figure}
    \centering
    \includegraphics[width=\linewidth]{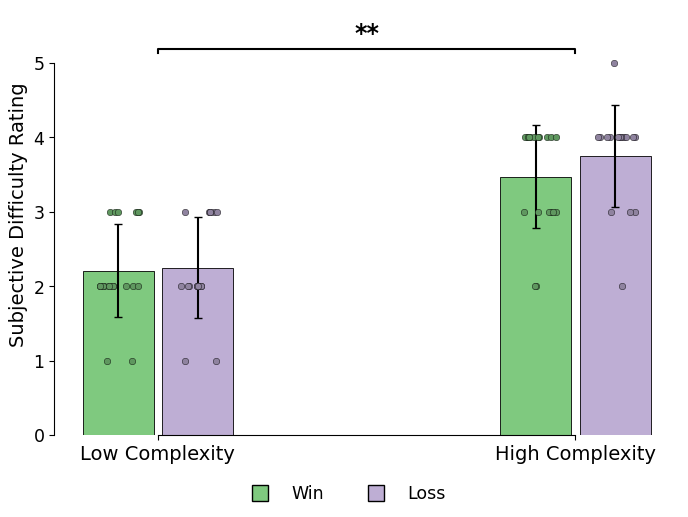}
    \caption{Subjective difficulty ratings across game sessions. Bars represent mean ratings ($\pm$ SD), separated by Low Complexity and High Complexity sessions, with performance groups shown as grouped bars (Win = high-performing; Loss = low-performing).
    Dots indicate individual participants’ ratings. Asterisks indicate significant difference between sessions (**$p < .01$).}
    \Description[Bar chart of complexity ratings by session and outcome.]
    {The figure presents self-reported difficulty ratings across two game sessions. 
    Ratings are higher in the high-complexity session compared to the low-complexity session. 
    Within each session, bars are grouped by outcome, showing separate values for win and loss participants. 
    Individual ratings are displayed as dots, and asterisks indicate a significant difference at $p < .01$.}
    \label{fig:difficulty}
\end{figure}
\begin{figure*}
    \centering
    \includegraphics[width=\textwidth]{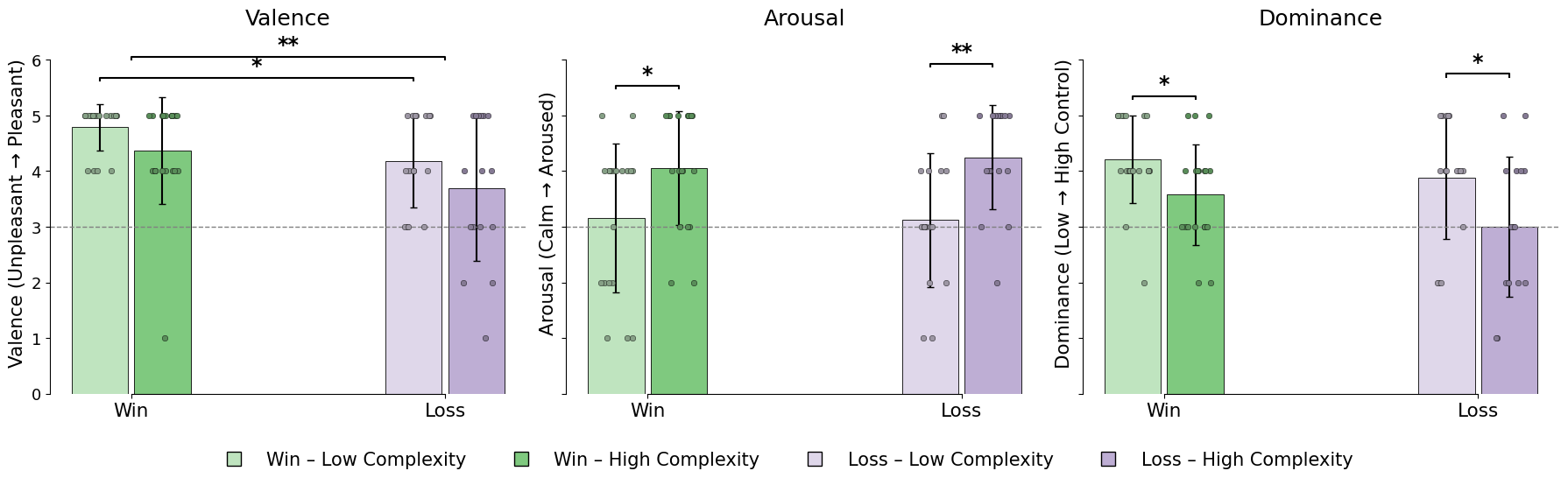}
    \caption{Subjective ratings on the SAM scales grouped by performance (Win = high-performing; Loss = low-performing).  
Bars show mean ($\pm$ SD), with dots indicating individual participants. 
Left: \textbf{Valence} (0 = unpleasant, 5 = pleasant). Middle: \textbf{Arousal} (0 = calm, 5 = aroused). Right: \textbf{Dominance} (0 = low control, 5 = high control). Asterisks indicate significant differences (**$p < .01$, *$p < .05$).}
  \label{fig:sam}
\Description[SAM ratings for valence, arousal, and dominance by complexity and outcome.]{The figure shows self-assessment manikin ratings across two sessions and outcome groups. 
For valence, participants in the win group reported higher pleasantness than those in the loss group. 
For arousal, ratings were significantly higher in the high-complexity session than in the low-complexity session for both performance groups. 
For dominance, perceived control was significantly lower in the high-complexity session than in the low-complexity session for both performance groups.  
Significant differences are marked with asterisks at *$p < .05$ and **$p < .01$.}
\end{figure*}
To examine perceived task demands across two sessions, subjective difficulty ratings were analysed across game complexity sessions using independent-samples \textit{t}-tests (Figure~\ref{fig:difficulty}). On average, difficulty ratings were $2.23 \pm 0.65$ in the Low Complexity session and $3.60 \pm 0.69$ in the High Complexity session across all participants. Ratings were significantly higher in the High Complexity session than in the Low Complexity session (\(p < .01\)). In addition, visual inspection showed that participants in the low-performing group reported more ratings at the upper end of the difficulty scale compared to the high-performing group. Within each complexity level, no significant differences were observed between performance groups (Low Complexity: high = $2.21 \pm 0.63$ vs.\ low = $2.25 \pm 0.68$; High Complexity: high = $3.47 \pm 0.70$ vs.\ low = $3.75 \pm 0.68$).

Subjective ratings on the SAM scales were analysed using independent samples \textit{t}-tests.
Tests examined (1) between-group differences (high- vs.\ low-performing participants) within each complexity condition; (2) within-group differences between complexity conditions for each performance group; and (3) overall between-group differences across complexity conditions. Results are summarised in Figure~\ref{fig:sam}.

Valence, ranging from unpleasant/negative to pleasant/positive, showed a significant performance effect across complexity levels, with the high-performing participants reporting higher valence ($4.58 \pm 0.76$) than the loss group ($3.94 \pm 1.11$), \textit{p} < .01. In the Low Complexity condition, the high-performing group reported more positive valence ($4.79 \pm 0.42$) than the low-performing group ($4.19 \pm 0.83$), \textit{p} < .05. 

Arousal, ranging from calm and relaxed to highly aroused, showed a significant effect of task complexity. In the high-performing group, arousal was significantly higher in the High Complexity condition ($4.05 \pm 1.03$) than in the Low Complexity condition ($3.16 \pm 1.34$), \textit{p} < .05. A similar effect was observed in the low-performing group (High Complexity: $4.25 \pm 0.93$; Low Complexity: $3.12 \pm 1.20$), \textit{p} < .01. 

Dominance, reflecting the degree of control (submissive to dominant), was consistently higher in the Low Complexity condition compared to the High Complexity condition across both performance groups. The high-performing group reported higher dominance in Low Complexity ($4.21 \pm 0.79$) than in High Complexity ($3.58 \pm 0.90$), \textit{p} < .05, and the low-performing group showed the same pattern (Low Complexity: $3.88 \pm 1.09$; High Complexity: $3.00 \pm 1.26$), \textit{p} < .05. Although dominance ratings were consistently higher for the high-performing group, no overall performance-group effect was observed (win: $3.89 \pm 0.89$; loss: $3.44 \pm 1.24$; \textit{p} = .088).

\subsection{Predictive Performance Analysis}
Predictive performance was assessed across ocular, cardiac, and fused modalities under a cross-subject evaluation. Classification outputs were obtained at the participant level. Class-wise outcomes and component contributions were then examined to characterise model errors and the role of key features.

\subsubsection{Cross-Subject Classification Performance}
\begin{figure}
    \centering
    \includegraphics[width=\linewidth]{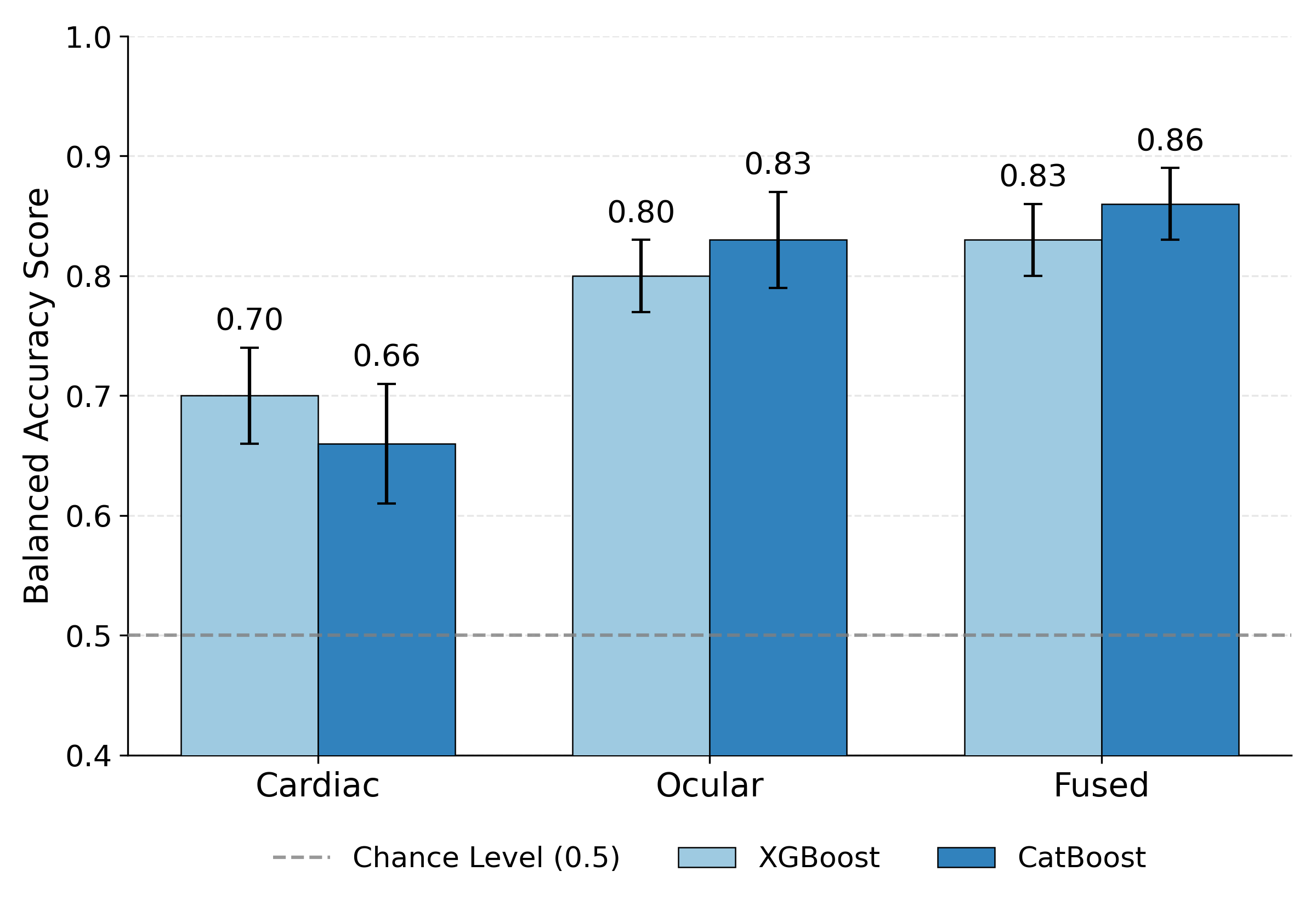}
    \caption{Cross-subject classification performance (balanced accuracy) comparison across modalities. The grouped bars illustrate the incremental predictive power from unimodal (cardiac, ocular) to multimodal (fused) approaches. CatBoost and XGBoost are shown as the top-performing classifiers}
    \Description[Grouped bar chart of model performance (balanced accuracy) comparison across modalities.]
    {The figure displays balanced accuracy values for CatBoost and XGBoost across three feature sets: cardiac, ocular, and fused. 
    Each feature set contains two adjacent bars corresponding to the two classifiers. 
    Fused models show the tallest bars overall, followed by ocular models, and cardiac models have the shortest bars. 
    CatBoost bars are consistently higher than XGBoost bars within each feature set.}
    \label{fig:placeholder}
\end{figure}

We evaluated three classifiers (CatBoost, XGBoost, Linear SVM) using a cross-subject LOSO protocol across the ocular, cardiac, and fused models. 
All results are based on the final selected configurations after model tuning and feature selection. Table~\ref{tab:performance} reports balanced accuracy, precision, recall, macro-F1, and MCC, while Figure~\ref{fig:placeholder} visually summarises the performance comparison of the tree-based models across modalities.
\begin{table*}
\centering
\caption{
Cross-subject LOSO performance across the ocular, cardiac, and fused feature sets for all models.
The table reports the best subject-level results obtained under LOSO, including balanced accuracy, macro-precision, macro-recall, macro-F1, and MCC.
}
\begin{tabular}{l l c c c c c}
\toprule
\textbf{Modality} & \textbf{Model} & \textbf{BAcc} & \textbf{Macro-Pr} & \textbf{Macro-Re} & \textbf{Macro-F1} & \textbf{MCC} \\
\midrule
Ocular   & CatBoost  & 0.83 & 0.83 & 0.82 & 0.83 & 0.66 \\
         & XGBoost   & 0.80 & 0.77 & 0.76 & 0.79 & 0.54 \\
         & Linear SVM & 0.57 & 0.79 & 0.56 & 0.48 & 0.27 \\
\midrule
Cardiac  & CatBoost  & 0.66 & 0.67 & 0.66 & 0.66 & 0.33 \\
         & XGBoost   & 0.70 & 0.72 & 0.70 & 0.70 & 0.42 \\ 
         & Linear SVM & 0.50 & 0.27 & 0.50 & 0.35 & 0.00 \\
\midrule
Fused    & CatBoost  & 0.86 & 0.89 & 0.84 & 0.87 & 0.72 \\ 
         & XGBoost   & 0.83 & 0.88 & 0.79 & 0.83 & 0.66 \\
         & Linear SVM & 0.50 & 0.27 & 0.50 & 0.35 & 0.00 \\
\bottomrule
\end{tabular}
\label{tab:performance}
\end{table*}

\begin{figure}
    \centering
    \includegraphics[width=\linewidth]{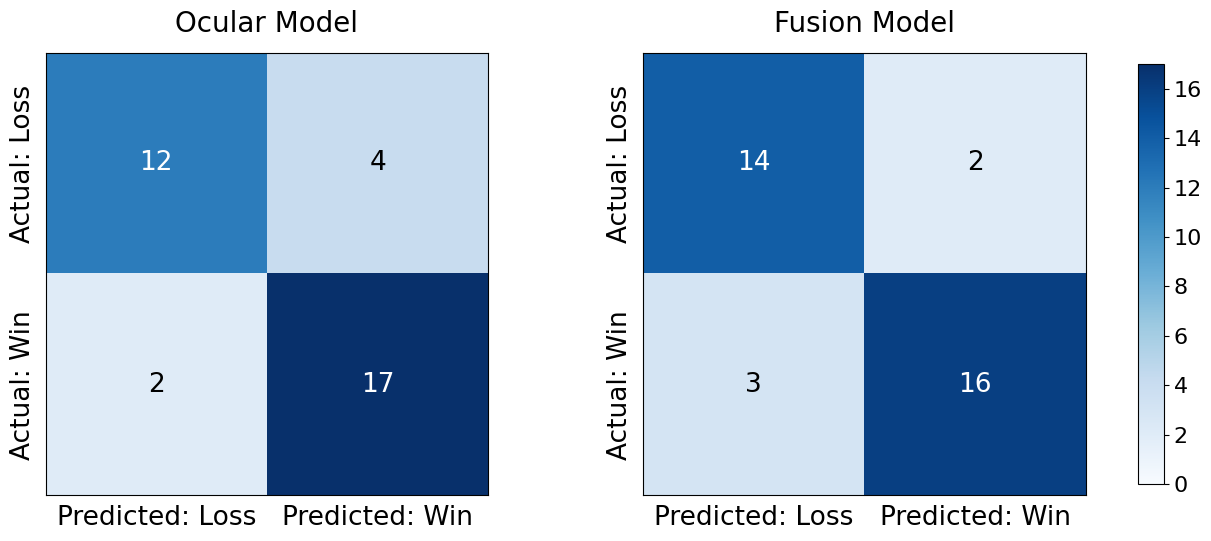}
    \caption{Confusion matrices for the ocular (left) and fused (right) models, selected as the top-performing classifiers from the cross-subject evaluation.}
     \Description[Confusion matrices for ocular vs fused classifiers.]
    {The figure shows two square confusion matrices, each with predicted labels on the x-axis and true labels on the y-axis.
    In both matrices, rows represent actual outcomes (Loss and Win), and columns represent predicted outcomes (Loss and Win).
    The left matrix corresponds to the ocular model: 12 correct Loss classifications in the top-left cell, 4 misclassified Loss samples in the top-right cell, 2 misclassified Win samples in the bottom-left cell, and 17 correct Win classifications in the bottom-right cell.
    The right matrix corresponds to the fused model: 14 correct Loss classifications, 2 misclassified Loss samples, 3 misclassified Win samples, and 16 correct Win classifications.
    Cell values are numerically indicated, with darker shading representing larger counts.}
    \label{fig:confusion_matrices}
\end{figure}
Across all models, the ocular feature set yielded the strongest unimodal performance.
CatBoost achieved a balanced accuracy of 0.83 (SE = 0.06) and an MCC of 0.66, and XGBoost performed similarly (BAcc = 0.80, MCC = 0.54).
Cardiac features provided weaker predictive value.
The best-performing cardiac model, XGBoost, achieved a balanced accuracy of 0.70 (SE = 0.07) with an MCC of 0.42.
The fused model produced the best overall cross-subject generalisation:
CatBoost reached a balanced accuracy of 0.86 (SE = 0.05) and an MCC of 0.72, while XGBoost showed comparable improvements (BAcc = 0.83, MCC = 0.66).
Across all three modalities, the Linear SVM consistently performed near chance level (BAcc $\approx 0.50$–$0.57$, MCC $\approx 0$–$0.27$), showing limited capacity to model the non-linear structure present in the physiological features.

Figure~\ref{fig:confusion_matrices} presents the confusion matrices for the ocular-only and fused models to detail classification accuracy across groups. The ocular model correctly identified 12 of the 16 actual 'Loss' cases, misclassifying 4 cases as 'Win' (false positives). In comparison, the fused model correctly identified 14 'Loss' cases, reducing the number of false positives to 2. Although the correct identification of 'Win' cases decreased slightly from 17 to 16, the integration of cardiac data produced a net increase in the accurate classification of the low-performing group.

\subsubsection{Ablation Study}
To evaluate the contribution of specific model components, we conducted ablation studies at two levels: (1) a module level analysis within the ocular modality to determine the predictive power of distinct feature groups, and (2) an architecture level analysis to assess the necessity of the consensus mechanism in the fusion model.

For the ocular model, we defined four key functional components:
(1) pupil-based measures,
(2) fixation statistics,
(3) saccade dynamics, and
(4) gaze allocation across AOIs.
We performed ablation tests by iteratively removing one component (including all associated features) and evaluating the resulting model. In each variant, only the targeted component was excluded while all remaining components were kept intact. All experiments followed the same cross-subject LOSO protocol, where each fold withheld one participant for testing. To ensure fair comparison, all model configurations, hyperparameters, and threshold-selection procedures were kept consistent with the full ocular model.
\begin{table}
\centering
\caption{Ablation of ocular feature modules, where one component is removed at a time. $\Delta$BAcc denotes the change in balanced accuracy relative to the full ocular model. The row \textbackslash\ corresponds to the full ocular model ($N=35$, LOSO).}
\label{tab:eye_catboost_ablation}
\begin{tabular}{@{}lcccc@{}}
\toprule
\textbf{Component Removed} & \textbf{BAcc} & \textbf{Macro F1} & \textbf{MCC} & $\Delta$\textbf{BAcc} \\
\midrule
\textbackslash & \textbf{0.83} & \textbf{0.83} & \textbf{0.66} & 0.00 \\
Pupil       & 0.83 & 0.83 & 0.66 & 0.00 \\
Fixation    & 0.83 & 0.83 & 0.66 & 0.00 \\
Saccade     & 0.69 & 0.68 & 0.37 & 0.14 \\
AOI Gaze    & 0.40 & 0.39 & -0.22 & 0.43 \\
\bottomrule
\end{tabular}
\end{table}
As shown in Table~\ref{tab:eye_catboost_ablation}, the ocular modality exhibited a clear internal structure. Some components provided redundant support, whereas others were essential for maintaining predictive reliability. Removing either the pupil features or fixation metrics produced no measurable degradation, with accuracy remaining at 0.83. In contrast, excluding saccade dynamics resulted in a substantial drop to 0.69. The largest degradation occurred when removing AOI-based gaze allocation, where accuracy fell to 40 percent. These results show that while basic ocular features contribute marginally, performance-critical information is carried primarily by saccadic behaviour and gaze allocation across task-relevant regions.

\begin{table}
  \centering
  \caption{Ablation of the consensus check in the stacking architecture, where one component is removed at a time. $\Delta$BAcc denotes the change in balanced accuracy relative to the full fusion architecture. The row \textbackslash\ corresponds to the full fusion architecture ($N=35$, LOSO).}
  \label{tab:fusion_catboost_ablation}
  \renewcommand{\arraystretch}{1.2}
  \setlength{\tabcolsep}{4pt}
  \begin{tabularx}{\linewidth}{>{\raggedright\arraybackslash}X c c c c}
    \toprule
    \textbf{Component Removed} & \textbf{BAcc} & \textbf{Macro F1} & \textbf{MCC} & $\Delta$\textbf{BAcc} \\
    \midrule
    \textbf{\textbackslash} & \textbf{0.86} & \textbf{0.86} & \textbf{0.72} & \textbf{0.00} \\
    Consensus Check & 0.77 & 0.77 & 0.54 & 0.09 \\
    \bottomrule
  \end{tabularx}
\end{table}

We also evaluated the contribution of the consensus mechanism within the full fusion architecture. We compared the full model with a variant that omitted the consensus check, where the meta-classifier was applied to all samples regardless of unimodal agreement. As presented in Table~\ref{tab:fusion_catboost_ablation}, removing the consensus step led to a drop in performance: balanced accuracy decreased from 0.86 to 0.77 ($\Delta$BAcc = 0.09), with macro-F1 and MCC dropping from 0.86 to 0.77 and from 0.72 to 0.54, respectively. These results indicate that applying the meta-classifier only to disagreement cases yields higher performance than applying it to all samples.

\subsection{Dynamics of Ocular and Cardiac Features across Performance Groups}
To interpret the predictive model, we examined the contribution of ocular and cardiac features using the feature-importance profiles of the best-performing model for each modality. Specifically, CatBoost was used to extract the importance of ocular features, and XGBoost was used for cardiac features. From these importance profiles, we identified the top-ranked features and selected a subset that aligned with established physiological and task-related interpretability. In total, five ocular features and two cardiac features were identified as the most relevant predictors in the unimodal models. Group differences in these features were analysed using independent-samples \textit{t}-tests or Mann--Whitney \textit{U}-tests, with the test choice determined by Shapiro–Wilk normality assessment. The full importance scores for all features are provided in Appendix~\ref{app:predictor_score} (Table~\ref{tab:eye_catboost_importance}).
\begin{figure}
    \centering
    \includegraphics[width=\linewidth]{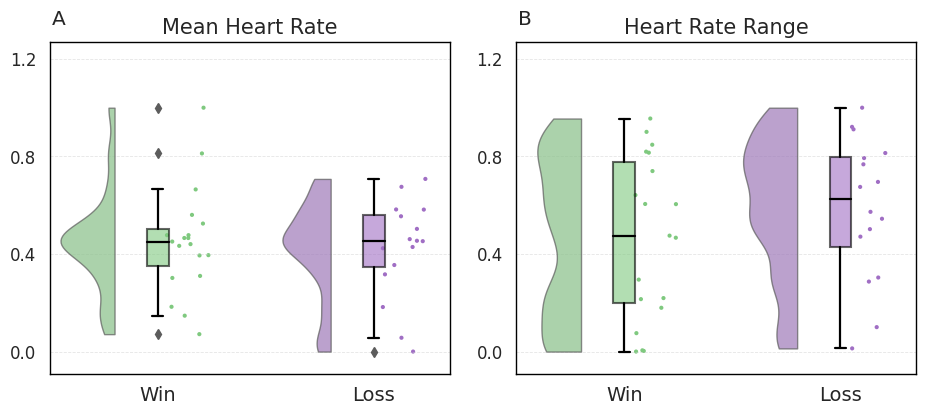}
    \caption{Raincloud plots of cardiovascular features comparing performance groups (Win = high-performing; Loss = low-performing). (A) Mean heart rate. (B) Heart rate range.
    For all panels, raincloud plots depict individual datapoints (dots), probability density (violin), and boxplots summarising the median and interquartile range.}
    \Description[Raincloud plots of cardiovascular features by performance group.]
    {The figure compares cardiovascular features between the win and loss groups.
    Panel A shows mean heart rate, and Panel B shows heart rate range.
    In all panels, individual participants are shown as dots, the distribution is shown as a violin representing probability density, and a boxplot summarises the median and interquartile range.}
    \label{fig:cardiac_feature}
\end{figure}
Figure~\ref{fig:cardiac_feature} illustrates cardiac indices. Mean heart rate (panel A) demonstrated a relatively compact interquartile range but a pronounced upper tail in the high-performing group compared with the low-performing group, although the difference did not reach statistical significance.
Heart rate range (panel B) exhibited greater dispersion, with the low-performing group showed an upward shift with a higher median, however, this difference was not statistically significant.
\begin{figure}
    \centering
    \includegraphics[width=\linewidth]{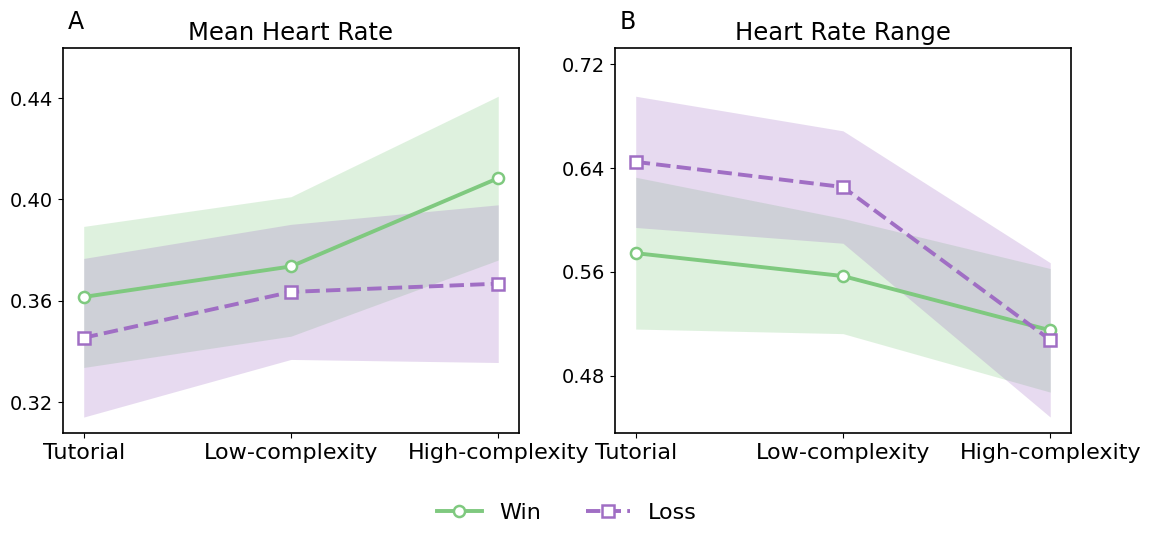}
    \caption{
    Dynamic trends of cardiac features across three phases (tutorial, low-complexity, high-complexity), comparing performance groups (Win = high-performing; Loss = low-performing).
    Lines show group means with shaded bands indicating $\pm$1 SEM.}
    \Description[Line plots of cardiac features across sessions by outcome group.]
    {The figure presents two line plots showing cardiac features across tutorial, low-complexity, and high-complexity sessions for win and loss groups.
    Solid lines indicate group means, with shaded bands representing variability within groups at plus or minus one standard error of the mean.
    Asterisks above the markers denote sessions where between-group differences are statistically significant, tested using either independent-samples \textit{t}-tests or Mann--Whitney \textit{U}-tests.
    Panel~A shows mean heart rate, and Panel~B shows heart rate range.}
    \label{fig:cardiac_trend}
\end{figure}
\begin{figure*}
    \centering
\begin{minipage}{0.84\textwidth}
\centering
\begin{subfigure}[b]{0.33\textwidth}
    \centering
    \includegraphics[width=\linewidth]{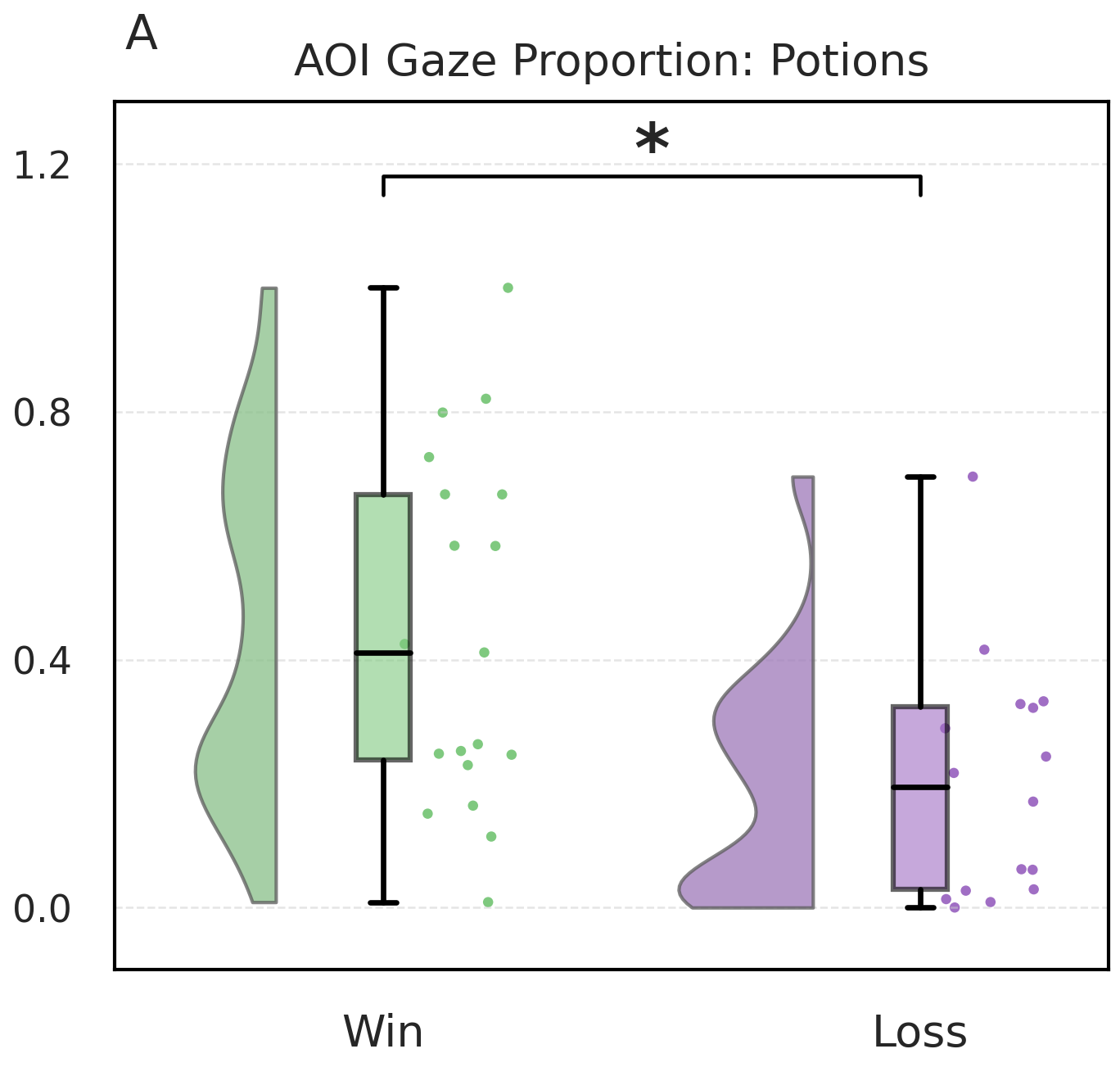}
\end{subfigure}
\begin{subfigure}[b]{0.33\textwidth}
    \centering
    \includegraphics[width=\linewidth]{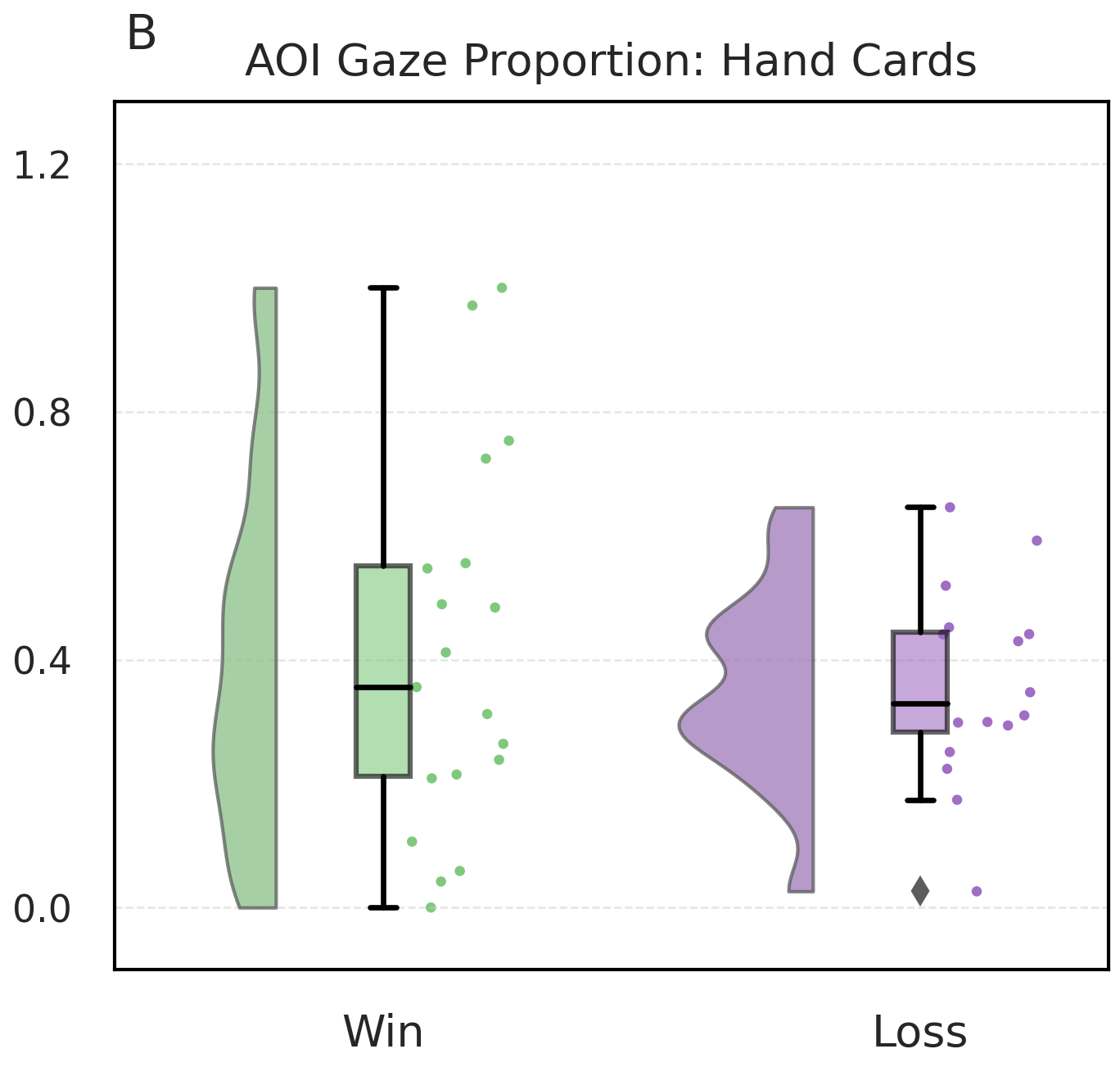}
\end{subfigure}

    \begin{subfigure}[b]{0.33\textwidth}
        \centering
        \includegraphics[width=\linewidth]{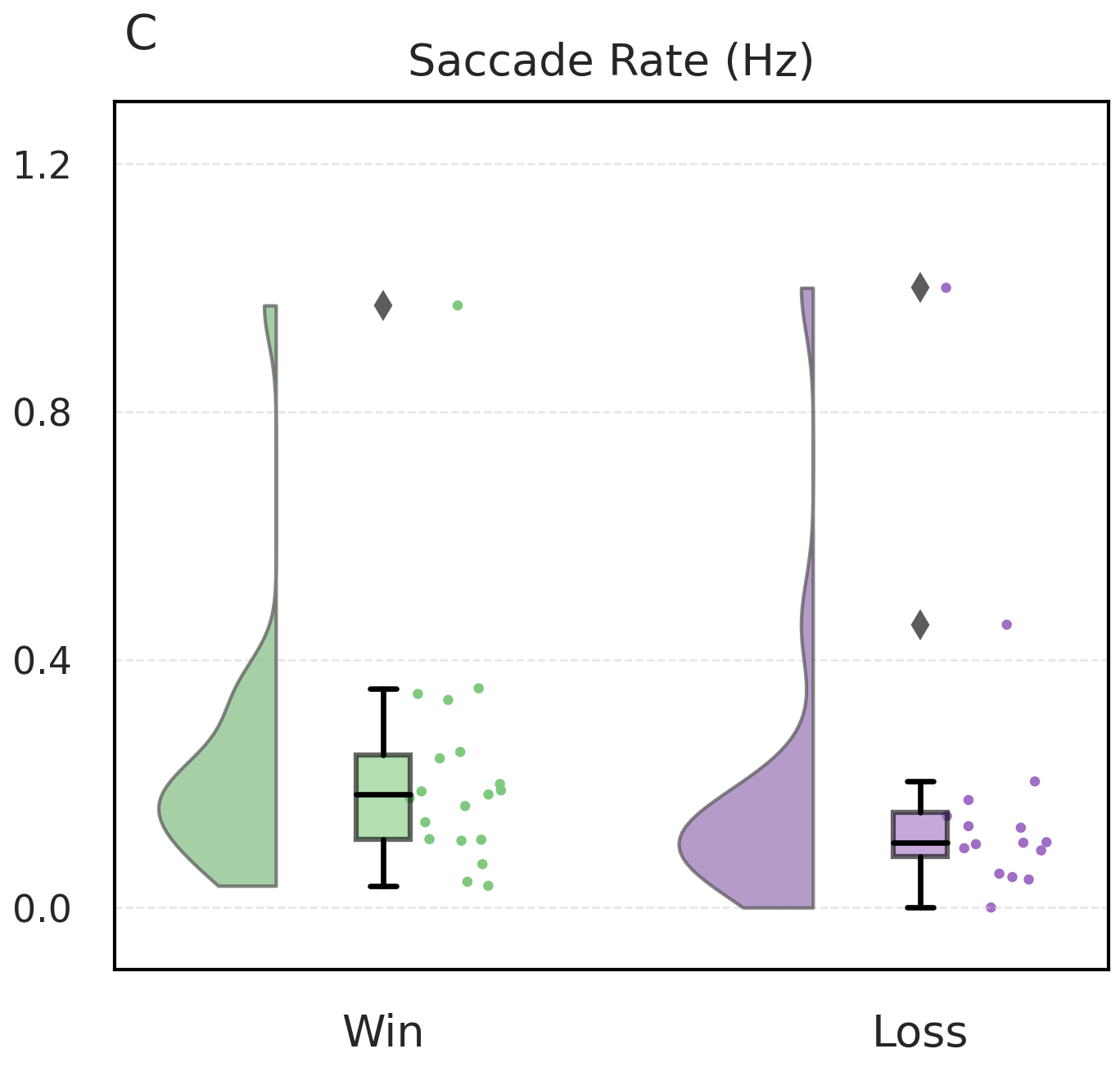}
    \end{subfigure}%
    \hfill
    \begin{subfigure}[b]{0.33\textwidth}
        \centering
        \includegraphics[width=\linewidth]{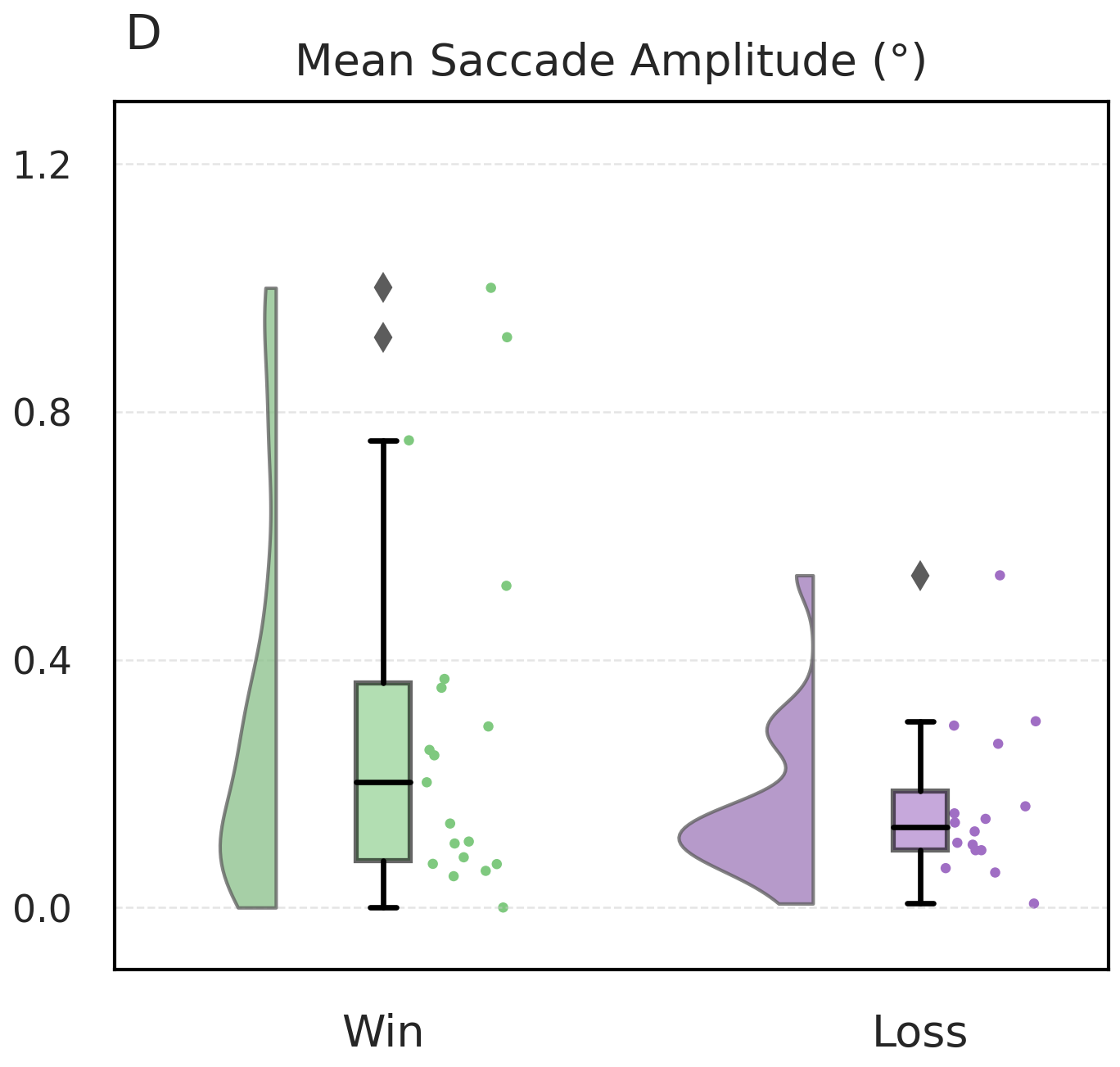}
    \end{subfigure}%
    \hfill
    \begin{subfigure}[b]{0.33\textwidth}
        \centering
        \includegraphics[width=\linewidth]{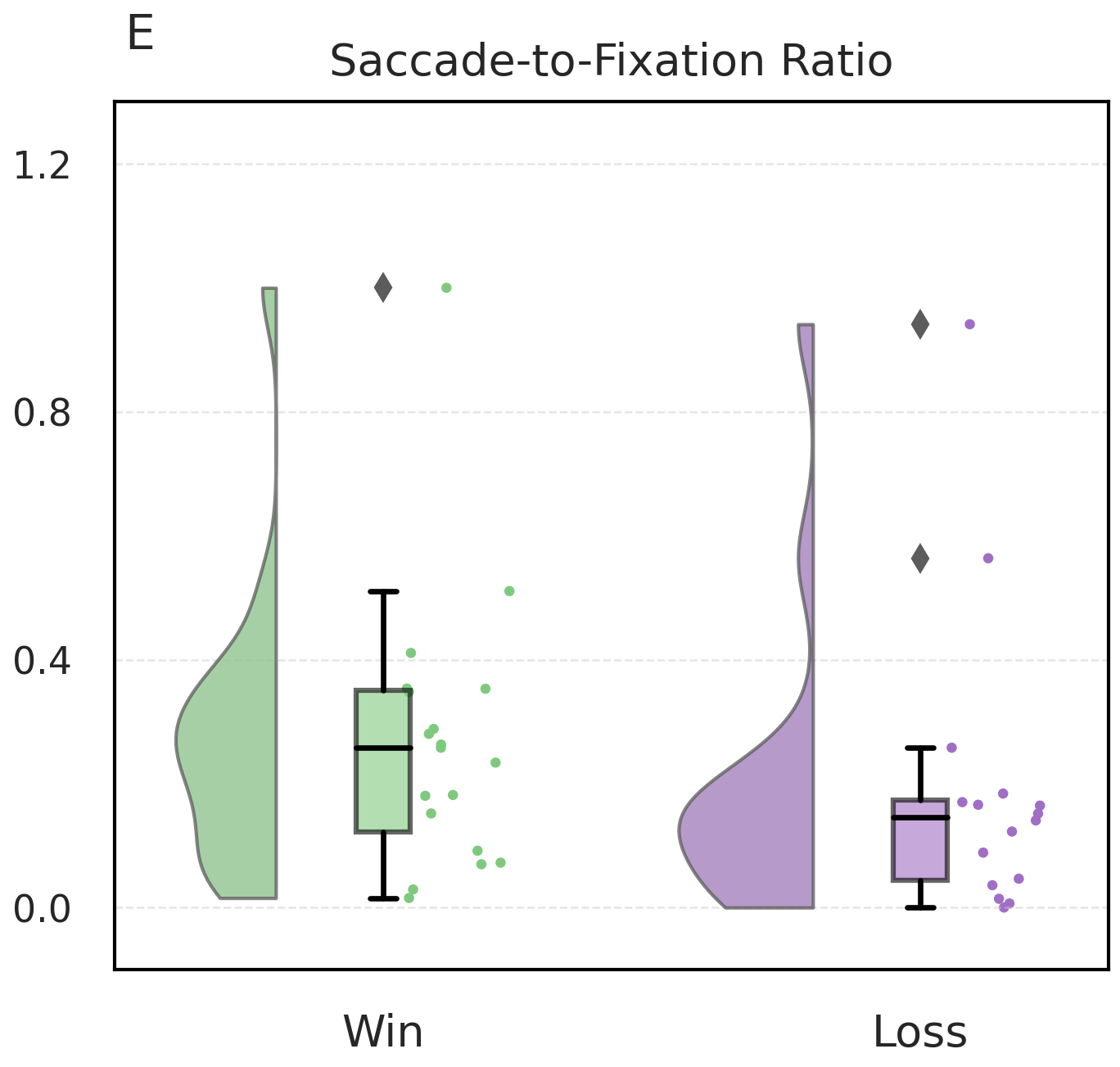}
    \end{subfigure}
\end{minipage}
    \vspace{0.2em}

    \includegraphics[width=0.3\textwidth]{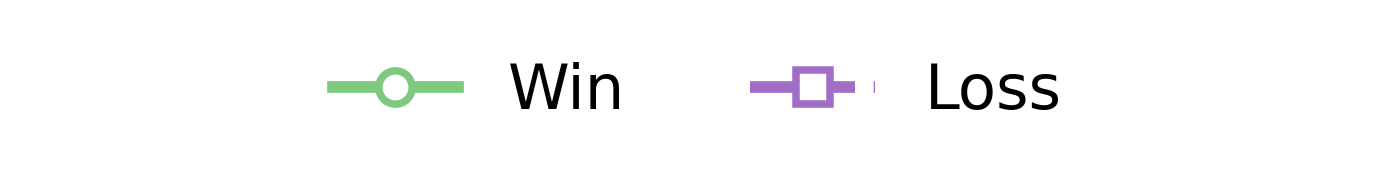}
    
    \caption{Raincloud plots of ocular features comparing performance groups (Win = high-performing; Loss = low-performing). (A) AOI gaze proportion: Potions. (B) AOI gaze proportion: Hand Cards. (C) Saccade rate (Hz). (D) Mean saccade amplitude (°). (E) Saccade-to-fixation ratio. For all panels, raincloud plots depict individual datapoints (dots), probability density (violin), and boxplots summarising the median and interquartile range. Asterisks indicate significant differences (*\textit{p} < .05).}
    \Description{
        Raincloud plots of ocular features comparing performance groups, with high-performing players labelled as Win and low-performing players labelled as Loss.
        The figure consists of five panels.
        Panel A shows AOI gaze proportion for Potions.
        Panel B shows AOI gaze proportion for Hand Cards.
        Panel C shows saccade rate measured in hertz.
        Panel D shows mean saccade amplitude measured in degrees.
        Panel E shows the saccade-to-fixation ratio.
        In all panels, individual participants are shown as dots, the distribution is shown as a violin representing probability density, and a boxplot summarises the median and interquartile range.
        Asterisks indicate statistically significant differences between groups at p less than 0.05.
}
    \label{fig:ocular_feature}
\end{figure*}

Figure~\ref{fig:cardiac_trend} presents the temporal trajectories of cardiac features across tutorial, low-complexity, and high-complexity sessions, separated by performance groups.
For mean heart rate (panel A), the high-performing group showed a steady increase from the tutorial to the high-complexity session. The low-performing group displayed a similar upward trajectory but remained lower at every stage. The separation between groups widened at higher complexity.
For heart rate range (panel B), both groups demonstrated a monotonic reduction across sessions. The low-performing group maintained larger heart rate ranges than the high-performing group at all stages, with the largest numerical separation in the tutorial phase. However, no session showed statistically reliable differences.

Figure~\ref{fig:ocular_feature} illustrates ocular indices.
(A) For AOI gaze proportion (potions), a significant between-group difference was observed (\textit{p} < .05). The high-performing group showed a higher median viewing rate with a broader spread towards higher values, while the low-performing group’s values were more concentrated at the lower end.
(B) For AOI gaze proportion (hand cards), no significant between-group difference was detected.
The high-performing group showed a slightly higher median viewing proportion with more dispersed samples across the mid-range, whereas the low-performing group exhibited a narrower distribution centred around lower gaze proportions.
(C) Mean saccade rate did not differ significantly between groups, but showed a trend towards higher values in the high-performing group (\textit{p} = 0.08). The high-performing group exhibited higher medians and longer upper tails, whereas the low-performing group remained centred at lower values.
(D) Mean saccade amplitude did not differ significantly between groups. The high-performing group showed a higher median and a slightly wider interquartile range, whereas the low-performing group had lower central tendencies and reduced spread.
(E) The saccade-to-fixation ratio did not reach statistical significance between groups (\textit{p} = 0.06), but showed a clear upward shift in the high-performing group. The high-performing group exhibited a higher median, with the interquartile range centred at higher values than the low-performing group.
\begin{figure*}
    \centering
\begin{minipage}{0.9\textwidth}
\centering

        \begin{subfigure}[b]{0.33\textwidth}
            \centering
            \includegraphics[width=\linewidth]{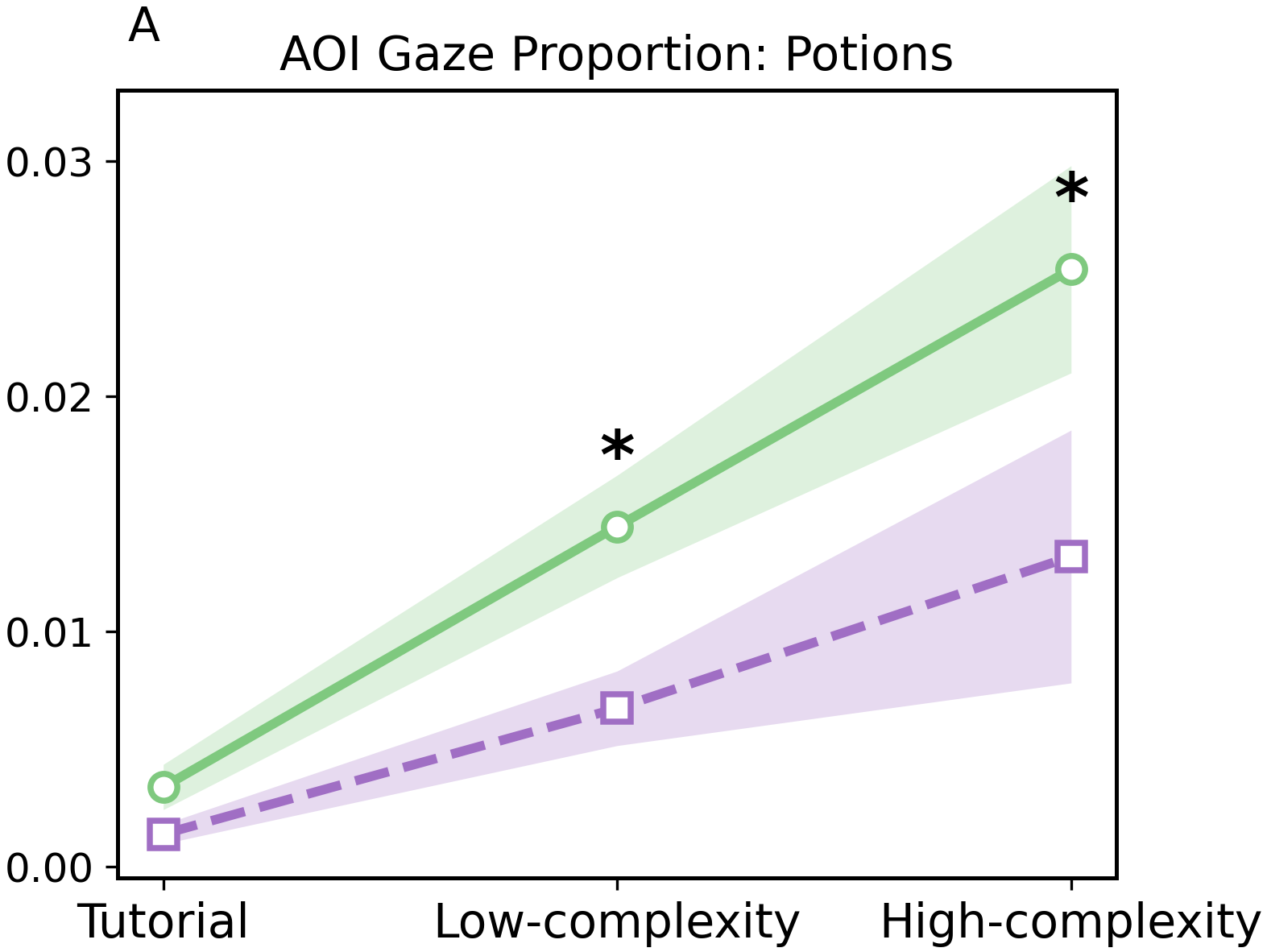}
        \end{subfigure}
        \begin{subfigure}[b]{0.33\textwidth}
            \centering
            \includegraphics[width=\linewidth]{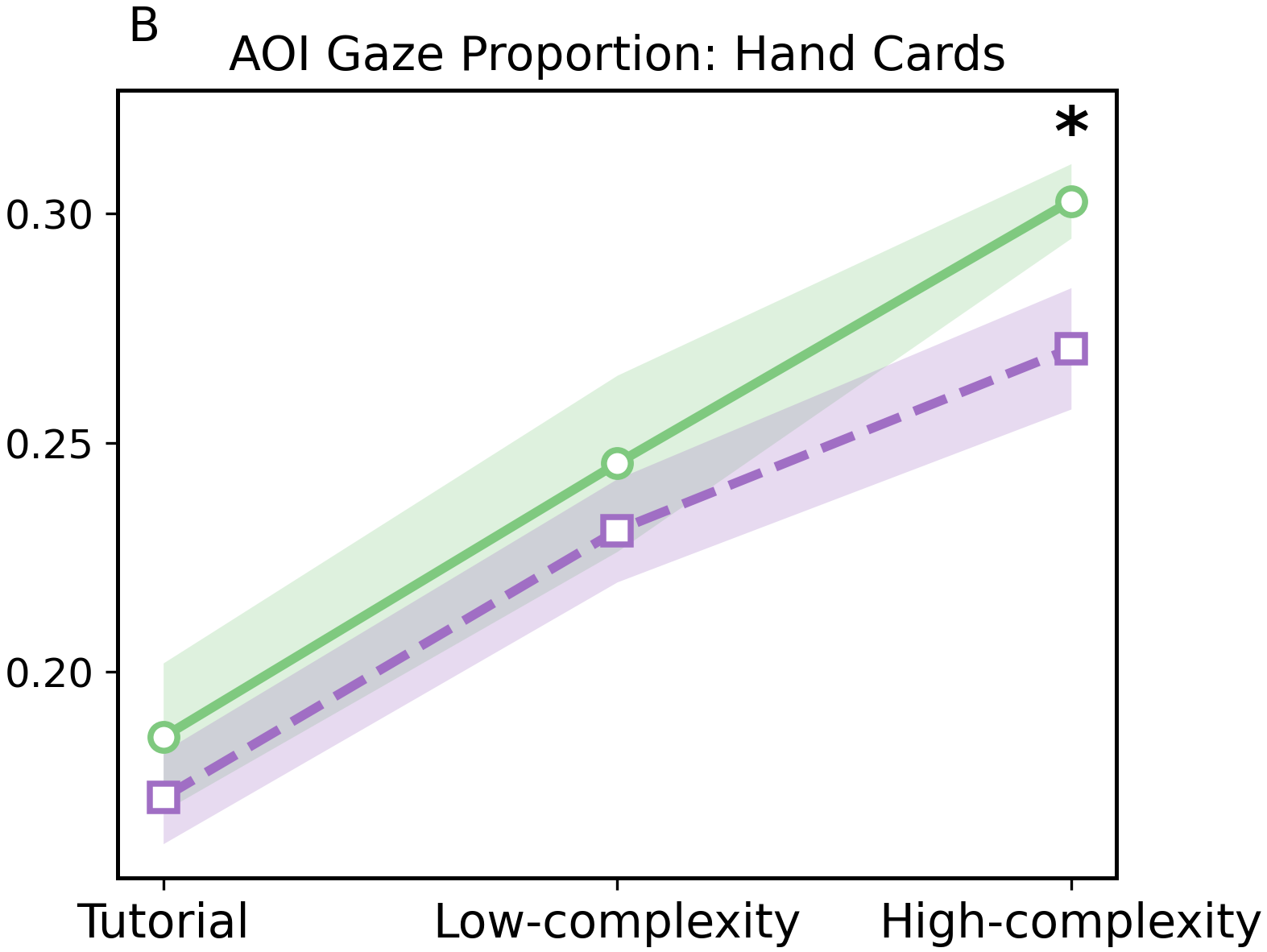}
        \end{subfigure}

    \vspace{1em}

    \begin{subfigure}[b]{0.33\textwidth}
        \centering
        \includegraphics[width=\linewidth]{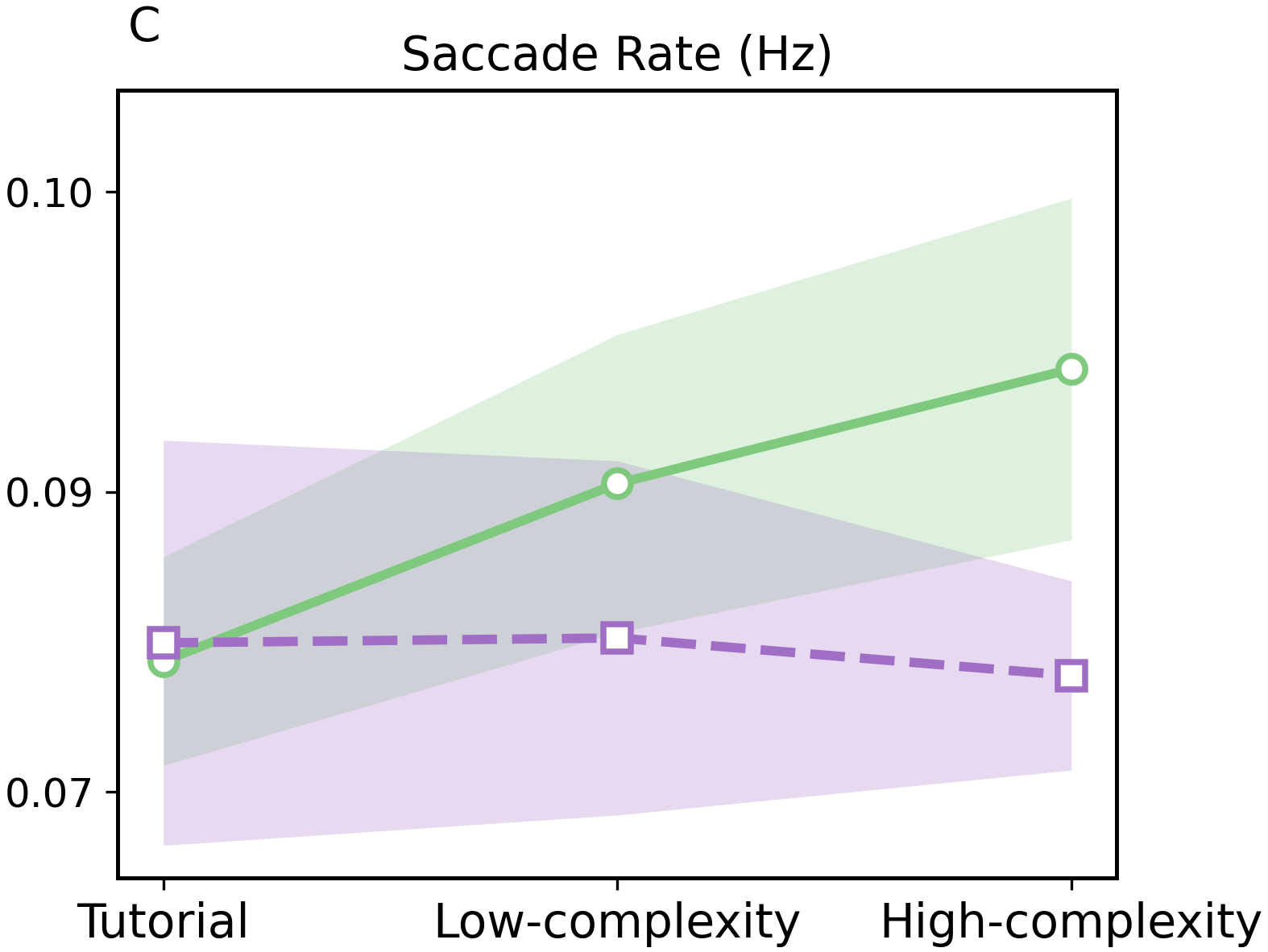}
    \end{subfigure}%
    \hfill
    \begin{subfigure}[b]{0.33\textwidth}
        \centering
        \includegraphics[width=\linewidth]{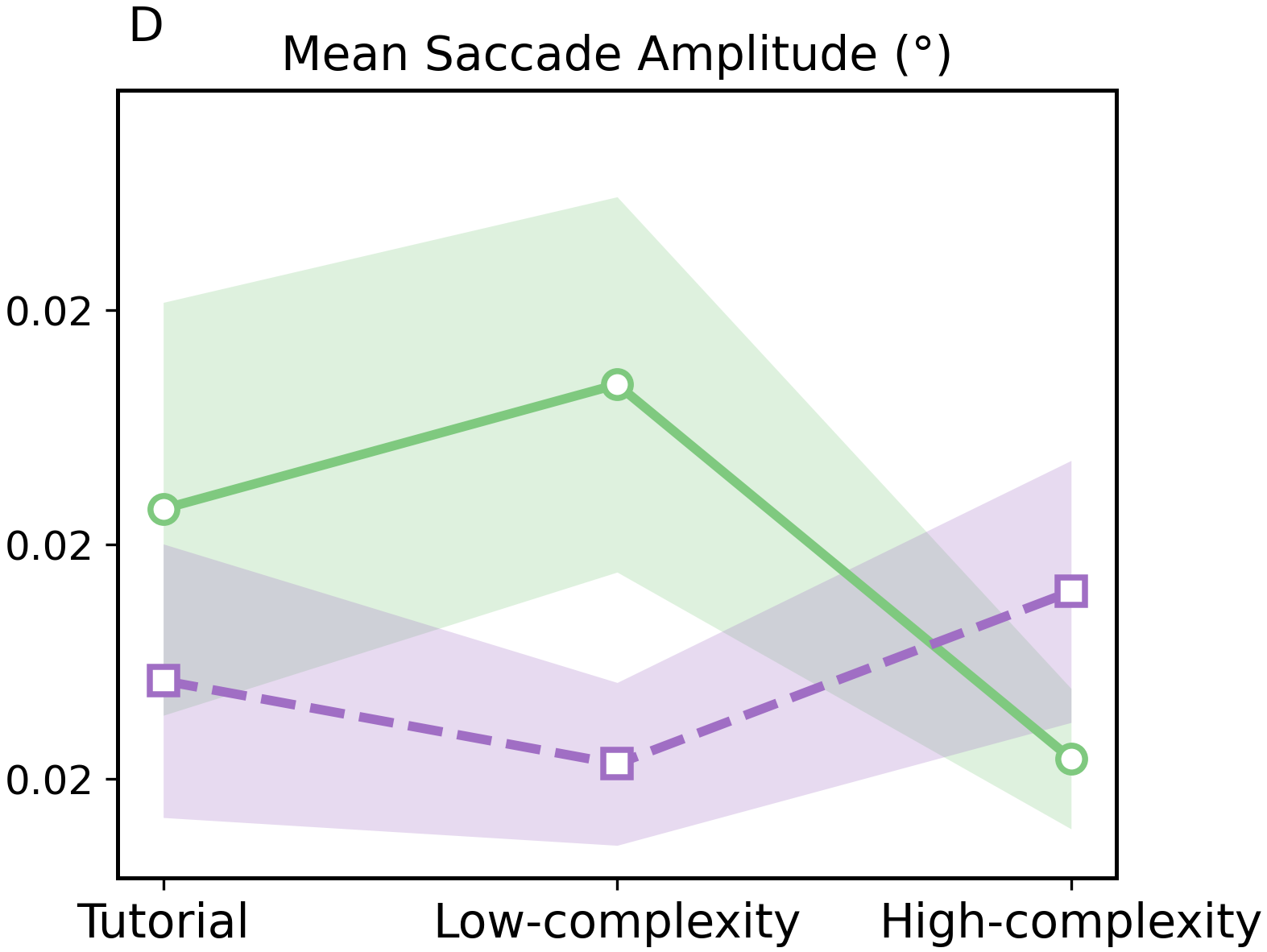}
    \end{subfigure}%
    \hfill
    \begin{subfigure}[b]{0.33\textwidth}
        \centering
        \includegraphics[width=\linewidth]{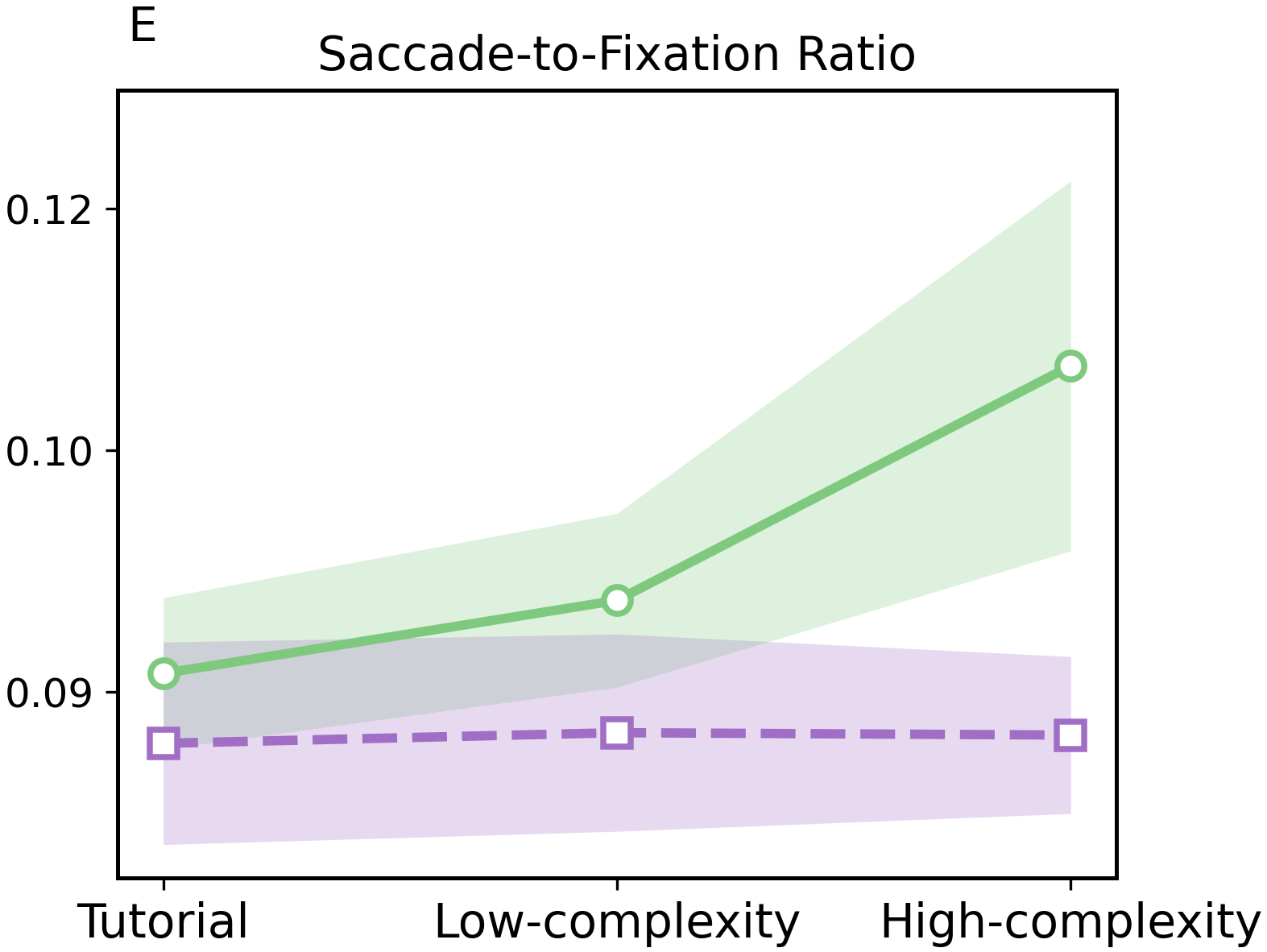}
    \end{subfigure}
\end{minipage}
    \vspace{0.2em}

    \includegraphics[width=0.3\textwidth]{Fig_Legend.png}

    \caption{
    Dynamic trends of ocular features across three phases (tutorial, low-complexity, high-complexity), comparing performance groups (Win = high-performing; Loss = low-performing).
    Lines show group means with shaded bands indicating $\pm$1 SEM. Asterisks indicate significant between-group differences at each session (*\textit{p} < .05).}
    \Description[Line plots of ocular indices across game sessions.]
    {The figure presents five line plots of ocular indices across three game phases: tutorial, low-complexity, and high-complexity. Panel A shows AOI gaze proportion for Potions. Panel B shows AOI gaze proportion for Hand Cards. Panel C shows saccade rate measured in hertz. Panel D shows mean saccade amplitude measured in degrees. Panel E shows the saccade-to-fixation ratio.
    Solid green lines show the win group and dashed purple lines show the loss group. Circles indicate win-group means, and squares indicate loss-group means.
    Asterisks mark statistically significant between-group differences at $p < .05$.}
    \label{fig:ocular_trend}
\end{figure*}

Figure~\ref{fig:ocular_trend} presents the temporal trajectories of ocular features.
(A) AOI gaze proportion (potions) increased progressively across sessions in both groups. The high-performing group exhibited consistently higher values, with significant between-group differences emerging from the low-complexity session onward (\textit{p} < .05). While both groups showed comparable baselines in the tutorial phase, the high-performing group demonstrated a steeper increase and reached the highest gaze proportion in the high-complexity session.
(B) AOI gaze proportion (hand cards) increased progressively across sessions for both groups. The two groups showed similar levels during the tutorial and low-complexity phases. A divergence emerged in the high-complexity session, where the high-performing group showed a significantly larger increase and achieved a higher proportion than the low-performing group (\textit{p} < .05).
(C) Saccade rate showed distinct patterns between groups. Both groups began at comparable levels in the tutorial phase, after which the high-performing group exhibited a steady increase, showing an upward trend relative to the low-performing group in the low-complexity session  (\textit{p} = 0.06). The low-performing group remained relatively stable and showed a slight decrease in the high-complexity session, resulting in increasing group divergence.
(D) Mean saccade amplitude displayed different trajectories between groups. The high-performing group increased from the tutorial to the low-complexity session, followed by a sharp decline in the high-complexity session. In contrast, the low-performing group exhibited an opposite trajectory, with an initial decrease followed by a subsequent increase. No statistically significant differences were found across sessions.
(E) The saccade-to-fixation ratio increased across sessions for the high-performing group, rising most prominently in the high-complexity phase. A trend towards higher values in the high-performing group was observed in the low-complexity session (\textit{p} = 0.08). By contrast, the low-performing group showed minimal change across stages, maintaining a nearly flat trajectory. 

\section{Discussion}
We evaluated the feasibility of using early physiological signals for performance prediction across task phases. In our study, participants completed two consecutive sessions of a deck-building task with escalating complexity (low and high), while ocular and cardiac activity and affective self-reports were recorded. We structure our discussion around three research questions: (1) assessing whether early ocular and cardiac signals are predictive of later performance outcomes as task complexity increases; (2) characterising the trajectories of physiological mechanisms reflected in key ocular and cardiac features that differentiate high- and low-performing groups across sessions; and (3) examining how subjective affective experiences performance outcomes diverge between these groups. We further consider broader implications of predictive ocular–cardiac signatures through two use cases that illustrate how these markers can be translated into human-state insights under demanding conditions.

\subsection{RQ1: Ocular Signals and Multimodal Fusion Enable Early Prediction of Performance under Escalating Complexity}
Addressing RQ1, our findings show that ocular features extracted from the early low-complexity session support the prospective prediction of later task performance. 
This finding extends performance-prediction research \citep{guo2025pptp, smerdov2023ai, heard2022predicting} by demonstrating that predictive information generalises across sessions in a continuously unfolding game environment. The early visual behaviour captured in the low-complexity phase exhibited capability across individuals to support the forecasting of later outcomes. Multimodal fusion further improved performance, indicating that cardiac information contributed complementary signals not captured by ocular features alone.

The performance of the ocular-only model reflects the close coupling between visual sampling demands and task progress in our environment.
Ablation analysis showed that its predictive value was driven by features describing where gaze was directed (AOI allocation) and how gaze shifted between elements (saccade dynamics). 
Their discriminative strength is consistent with prior work linking visual sampling behaviour to effective performance \citep{brams2019relationship, dogusoy2014cognitive, ooms2014study}.
In addition, although the specific AOIs were derived from our game interface, gaze allocation to task-relevant regions has been widely used as a performance-related indicator in visually demanding interactions \citep{fogarty2023eye, ooms2014study, li2023using}. 
In our task, the interface is visually stable yet requires active information search given the strategic nature of the game. Accordingly, these predictors likely captured patterns of how gaze was allocated and distributed associated with performance outcomes.
This suggests that ocular measures provide useful indicators for forecasting task performance, rather than merely describing viewing preferences.

On the other hand, pupillary features contributed less to early prediction.
Although pupil diameter is a well-established indicator of arousal \citep{marshall2007eye, hayes2016pupil} and cognitive control \citep{van2018pupil}, its utility in the present task was likely constrained by dense, rapidly evolving in-game events. Prior work shows that pupil diameter increases during exploratory states and decreases during exploitation states \citep{hayes2016pupil}. Participants may have experienced frequent event-evoked dilations and constrictions, producing oscillatory pupillary responses that weakened longer-term distinctions related to performance. In addition, frequent luminance variations within a visually dynamic game screen may have introduced light-evoked changes that further reduced discriminative power \citep{beatty2000pupillary}.

A parallel pattern emerged in the cardiac modality. Mean HR and HR range were the most predictive features, whereas HRV measures were weaker indicators in our task. Although HRV is typically associated with task-related regulation \citep{Thayer2009, colzato2018variable} and has shown predictive value during high-intensity game events \citep{long2023heart}, it primarily reflects short-term autonomic modulation. In contrast, our turn-based task required sustained engagement over longer intervals rather than rapid, event-driven reactions.
In this setting, overall levels of cardiac activation (HR) showed a stronger association with later performance, suggesting that tonic activation may offer more predictive value for cross-session prediction than momentary autonomic fluctuations.

Despite the lower unimodal accuracy of cardiac signals, the fused model achieved a more balanced class-wise performance by improving the identification of low-performing participants.
This suggests that cardiac features encode discriminative information that is not fully captured in the ocular features and becomes useful particularly when the ocular classifier is uncertain.
Additionally, the complementary value is localised to instances where the unimodal models diverge.
As shown in the ablation analysis, training the meta-classifier on all samples caused the optimisation landscape to be dominated by majority agreement cases, which limited learning of decision rules from the informative disagreement region where cardiac signals add value.
By restricting training to disagreement samples, the fused model was forced to learn from instances where the two modalities diverged, enabling it to exploit the complementary strength of cardiac information. This targeted integration produced the higher overall performance observed in the fused model.

\subsection{RQ2: Ocular and Cardiac Signatures Reveal Divergent Visual Behaviours and Arousal Responses Across Performance Groups}
Addressing RQ2, we identified two physiological patterns differentiating the performance groups across sessions:  
(1) how players visually accessed task-relevant information, reflected in saccade behaviour and gaze allocation; and
(2) how physiological activation changed with increasing demands, reflected in heart-rate responses.

Gaze behaviour reflected clear differences in how players accessed task-relevant information.
The high-performing group consistently allocated more gaze to strategically important interface elements throughout the task (potions and hand cards). Notably, the potion icons attracted sustained attention despite being located in peripheral regions. Such prioritisation has been widely associated with skilled viewing behaviour in visually demanding tasks \citep{fogarty2023eye, li2023using}. This aligns with the information-reduction perspective \citep{Haider1999}, which proposes that experienced performers concentrate processing on regions that are most useful for task progress. In our data, this translated into effective information access being established early and maintained throughout the task by the high-performing group.

Saccadic measures further revealed differences in how gaze shifts responded to increasing complexity.
Early in the task, the high-performing group displayed both higher saccade rates and larger amplitudes, indicating that their gaze sampled more on-screen elements at this stage \citep{phillips2008dependence}. Interestingly, as complexity escalated, their amplitudes decreased while rates remained high, indicating a shift towards faster checks of relevant regions rather than broad scanning. 
This pattern reflects a transition from exploration to more compact sampling, consistent with task-driven viewing behaviours among skilled users reported in prior work \citep{Hayhoe2005, Land2006, lu2021eyes}.
This change was accompanied by a steady increase in the saccade–to–fixation ratio, showing that more gaze time remained devoted to sampling multiple elements as the task complexity increased \citep{salvucci2000identifying}.
In contrast, the low-performing group exhibited little change in sampling behaviour: scanning remained less active overall, and their saccade–to–fixation ratio stayed stable across sessions, indicating limited adjustment in how gaze time was distributed between movement and inspection. Notably, their amplitude increased only when demands intensified — a trend that moved in the opposite direction to the high-performing group. The late increase in amplitude may reflect the need to scan more widely distributed information once demands peaked, suggesting that the timing at which useful information was accessed became critical for task performance.

Heart rate reflected the extent to which task-evoked effort was sustained as demands increased.
Within our study, HR rose for both groups when complexity escalated, a pattern consistent with prior observations that higher task demands are accompanied by elevated heart rate \citep{alshanskaia2024heart, jervcic2020modeling}. 
Differences emerged most clearly in the high-complexity phase: the high-performing group showed a continued rise in HR, whereas the low-performing group plateaued.
Prior work indicates that HR tends to increase when a task is experienced as demanding yet still attainable, whereas it stabilises when regulatory capacity is insufficient to meet rising demands \citep{richter2016three, richter2008task}. This suggests that maintaining elevated activation into the high-complexity phase appears more strongly associated with successful performance than initial reactivity alone, whereas stabilisation at a lower level coincided with poorer outcomes.

HR range further distinguished the groups. The low-performing group consistently exhibited larger HR ranges, indicating more variable cardiac activation. Such fluctuation suggests that their physiological state may have been more reactive to discrete in-game events (e.g., loss of HP or shifts in battle pace) \citep{drachen2010correlation}. 
These findings together reveal a non-linear cardiac response to escalating task complexity. In other words, what differentiated performance was not how high heart rate rose, but how reliably cardiac activation was maintained as demands escalated. Stable physiological mobilisation during the demanding phase appeared more consequential for eventual outcomes than initial reactivity alone.

Across ocular and cardiac channels, two hallmarks characterised successful performance in our study: (1) sustained visual attention towards task-relevant areas, accompanied by timely gaze shifts as complexity increased, and (2) stable cardiac activation when demands intensified.
Participants who showed both continued to access useful information and maintained the activation needed to act upon it under high demands, whereas others either did not prioritise relevant information sufficiently or exhibited less stable activation in the most demanding phase. These physiological patterns provide interpretable indicators of how information was accessed and how effort was sustained across escalating task demands.

\subsection{RQ3: Divergent Affective Trajectories and Sense of Control Across Performance Groups}
Addressing RQ3, we found participants’ affective responses to escalating task demands diverged by performance group, particularly in valence and dominance, despite similar increases in arousal and perceived difficulty. Across high- and low-performing groups, participants consistently reported greater perceived difficulty in the high-complexity stage, confirming that task complexity was effectively manipulated by the natural game progression, as reflected in the rise of arousal during the high-complexity session in both groups. Yet their affective trajectories diverged. Among high-performing participants, elevated arousal was accompanied by higher positive valence, a pattern consistent with Self-Determination Theory and flow theory \citep{sweetser2005gameflow, ryan2017sdt}, where arousal induced by optimal challenge sustains positive affect, provided that competence is preserved. Notably, even in the early stage, low-performing participants already reported more negative emotions (see Figure~\ref{fig:sam}), suggesting that limited familiarity or weaker perceived control of the game mechanics caused confusion or disengagement from the outset.

This lack of differentiation in overall affective arousal provides interpretative context for our prediction model results: physiological features that are associated with general arousal level (e.g., HR and Pupil) show limited discriminative power for performance prediction, as high arousal was a common state across participants in our context. At the same time, low-performing participants showed a stronger increase in arousal from low to high complexity, indicating greater sensitivity to task demand escalation. These patterns suggest that the way arousal changes with task demands may be more informative for distinguishing performance trajectories than the absolute arousal state itself.

Dominance trajectories also diverged between high- and low-performing groups. As complexity rose, reported dominance declined for all participants, but the high-performing group reported less erosion of dominance. For the low-performing group, the same failure situation led to divergent dominance experiences, as evidenced by the more polarised distribution. This variability suggests that low-performing participants did not interpret the failure state uniformly: while some retained a sense of agency, others experienced a marked collapse of control \citep{bopp2016negative, kapur2016examining, juul2013art}. Dominance therefore complements arousal by characterising how players construe difficulty, rather than how intensely they react to it. This affective dimension reveals distinct psychological pathways into failure that are not evident from physiological responses alone.

\subsection{Broader Implications: From predictive signatures to Human-State Insights}
Within a single game context, the observed ocular and cardiac patterns demonstrate the potential of physiological patterns to inform interpretable characterisations of human state under increasing task demands.
Our model shows that early ocular–cardiac patterns can forecast later performance under escalating complexity: fewer saccades and a narrower scan range (Fig.~\ref{fig:ocular_trend} C, D, E), and reduced gaze distribution over the critical AOI (Fig.~\ref{fig:ocular_trend} A, B) were characteristic of at-risk trajectories, whereas high-performing participants exhibited broader and more frequent saccades and allocated gaze more consistently toward task-relevant elements.
Cardiac signals further showed that high-performing participants maintained higher HR throughout (Fig.~\ref{fig:cardiac_trend}A), whereas the low-performing group exhibited a consistently larger HR range (Fig.~\ref{fig:cardiac_trend}B). These multimodal signatures provide early markers of visual strategies and physiological activation that precede measurable differences in performance.

\textbf{Use case 1 – Early signatures of limited visual sampling.}
In the initial phase of tasks, saccadic and gaze allocation patterns associated with proactive visual behaviour can serve as measurable indicators of emerging strain. 
Lower saccade rates, narrow saccade amplitudes, and limited gaze on critical regions indicate a lack of proactive information sampling, where access to task-relevant cues becomes more delayed or less frequent.
These signatures delineate a window in which suboptimal allocation of visual attention and information scanning emerges before observable differences in performance, particularly in visually intensive tasks with a relatively stable interface structure.
Early detection of these patterns could inform design approaches to scaffolding or information guidance in contexts where sustaining focus on task-relevant elements is critical, for instance redirecting visual attention towards elements that most strongly support task progress.

\textbf{Use case 2 – Phase-transition markers.} 
At natural transitions in task structure, such as shifts from low to high workload, cardiac measures provide early indicators of how well physiological activation is being sustained.
A plateau in HR across a transition suggests that mobilisation is not increasing in line with escalating demands, while a larger HR range indicates more reactive, event-driven fluctuations.
These features reveal moments when activation becomes less stable under pressure, offering a physiological lens on when individuals may be more vulnerable to errors.
These markers could support strategic pauses or checkpoint-guided interventions in domains such as training, education, and safety-critical operations, where phase transitions are central to task structure.

\subsection{Limitations and Future Work}
Our study was conducted in a single game, \textit{Slay the Spire}, which provided a progressively challenging and ecologically valid environment for testing predictive physiology. However, this setting emphasises individual decision-making and turn-based strategy. 
This limitation is particularly relevant for ocular measures, which are known to be highly sensitive to task structure and visual layout. Consequently, the predictive value of eye-movement features observed in this study should be interpreted as context-specific.
Future work should evaluate whether the same multimodal patterns emerge in broader interactive contexts that place different demands on users, such as sustained collaboration or safety-critical training tasks. Examining these varied settings would help establish the boundary conditions of physiological forecasting and determine how robustly ocular and cardiac signals generalise across domains.

A second limitation concerns the heterogeneity observed within the loss group. Performance outcome (high-performing vs. low-performing) served as the most robust and interpretable indicator of performance, yet our analysis of self-reported dominance showed marked variation among participants. Some retained a sense of control despite losing, whereas others experienced a complete collapse of agency. This suggests that the binary outcome may obscure distinct failure pathways. Future research should distinguish subtypes of low-performing participants, for example by clustering physiological and experiential profiles within the low-performing group and incorporating additional evaluative elements, thereby capturing how different breakdown trajectories manifest in both signals and experience.

A third limitation concerns the cardiac data acquisition. We utilised a wrist-worn PPG device sampling at 64 Hz, selected for its lightweight design and minimal disruption during gameplay. While this sampling rate is sufficient for heart-rate estimation, it lacks the temporal resolution required for reliable short-term HRV computation, where millisecond-level precision is necessary for capturing beat-to-beat variation. Although the inter-beat intervals were stabilised through interpolation, the HRV-derived features did not improve predictive performance. It therefore remains an open question whether higher-fidelity sensors would yield more informative HRV measures in this task context. As HRV is the most informative component of cardiac activity for inferring autonomic regulation, future work would benefit from incorporating higher-fidelity sensors or ECG-based systems to obtain more reliable HRV and allow a clearer assessment of its predictive value.

A fourth limitation concerns the absence of a pre-task physiological baseline. The study began with the gameplay sequence; although within-subject z-score normalisation reduced intra-individual variability, this procedure cannot recover information about participants’ tonic arousal or pre-existing anxiety levels prior to task onset. These baseline traits influence both pupil-linked arousal and autonomic cardiac reactivity and therefore impose an unknown offset on the subsequent task-evoked responses. Without a dedicated resting phase, it is difficult to determine whether observed physiological differences reflect genuine task-induced regulation or underlying stable traits. Future protocols incorporating a controlled resting baseline would enable clearer separation of task-evoked dynamics from individual physiological predispositions and improve interpretability of anticipatory physiological markers.

Finally, our modelling was validated in a relatively small participant cohort. Given the exploratory nature of this study, future work with larger and more diverse samples will be important for assessing the generalisability of current findings.

\section{Conclusions}
This study examined whether early ocular and cardiac signals can prospectively predict later performance outcomes as task complexity increases in a game environment. A within-subject experiment with 35 participants was conducted, combining ocular and cardiac measurements with self-reported affective ratings. We evaluated the predictive power of ocular, cardiac, and decision-level fused models under a LOSO cross-subject validation protocol. Beyond prediction, we analysed group-level feature trends across complexity phases and subjective affective reports to contextualise performance differences in relation to physiological dynamics and subjective appraisal.

In response to RQ1, ocular features and the fused model effectively predicted later performance, whereas the cardiac model alone was weaker. Across modalities, gaze distribution, saccade behaviour, and heart rate were the main predictive features. 
For RQ2, high-performing participants maintained targeted gaze on task-relevant elements, timely adapted their visual sampling as demands increased, and sustained more stable cardiac activation into the high-complexity phase, indicating distinct visual behaviour and autonomic mobilisation under pressure.
For RQ3, the high-performing group reported more positive affective experience, while the low-performing group displayed a wider spread in their reported sense of control.

In the future, we will extend this work to examine whether the physiological signatures observed in the present task generalise to broader interactive contexts, particularly those involving complex task dynamics and phase transitions. Further research will also explore heterogeneity within low-performing participants by identifying subtypes of failure trajectories through clustering of physiological and experiential profiles. In addition, using higher-sampling-rate sensors for HRV estimation would provide more reliable autonomic markers and allow a more precise characterisation of cardiac contributions to performance prediction. By addressing these directions, future work can extend beyond the current task setting and build multimodal indicators that support predictive systems capable of anticipating performance breakdowns, while advancing quantitative insight into how user states evolve under escalating task complexity.

\bibliographystyle{ACM-Reference-Format}
\bibliography{yufei}

\appendix

\section{Model Hyperparameter Configuration}
\label{app:hyperparams}
Table~\ref{tab:model_hyperparams} summarises the hyperparameters used for the CatBoost and XGBoost classifiers in the cross subject evaluation. The configurations were selected to provide stable training while limiting overfitting and keeping training time manageable.

\begin{table}
\centering
\caption{Model hyperparameters for SVM, CatBoost and XGBoost.}
\label{tab:model_hyperparams}
\begin{tabular}{ll}
\toprule
\textbf{Model} & \textbf{Hyperparameters} \\
\midrule

\textbf{Linear SVM} &
\begin{tabular}[c]{@{}l@{}}%
C = 1.0 \\
class\_weight = \{0: 1.0, 1: float(spw)\} \\
random\_state = 42 \\
\end{tabular}
\\
\midrule

\textbf{CatBoost} &
\begin{tabular}[c]{@{}l@{}}%
iterations = 600 \\
depth = 6 \\
learning\_rate = 0.05 \\
l2\_leaf\_reg = 3.0 \\
loss\_function = Logloss \\
eval\_metric = Logloss \\
verbose = False \\
random\_state = 42 \\
class\_weights = [1.0, max(1.0, spw)] \\
\end{tabular}
\\
\midrule

\textbf{XGBoost} &
\begin{tabular}[c]{@{}l@{}}%
n\_estimators = 400 \\
max\_depth = 5 \\
learning\_rate = 0.05 \\
subsample = 0.8 \\
colsample\_bytree = 0.8 \\
min\_child\_weight = 5 \\
reg\_lambda = 1.5 \\
reg\_alpha = 0.0 \\
gamma = 0.2 \\
objective = binary:logistic \\
class\_weights = [1.0, max(1.0, spw)]\\
tree\_method = hist \\
random\_state = 42 \\
eval\_metric = logloss \\
\end{tabular}
\\
\midrule
\makecell[l]{\textbf{XGBoost}\\\textbf{(Fusion Model)}} &
\begin{tabular}[c]{@{}l@{}}%
n\_estimators = 10 \\ 
max\_depth = 2 \\   
learning\_rate = 0.08 \\
subsample = 1.0 \\
colsample\_bytree = 1.0 \\
min\_child\_weight = 1.0 \\
reg\_lambda = 1.0 \\
objective = binary:logistic \\
scale\_pos\_weight = float(spw) \\
eval\_metric = logloss \\
n\_jobs = 1 \\
tree\_method = hist \\
random\_state = 42 \\
\end{tabular}
\\
\bottomrule
\end{tabular}
\end{table}

\section{Feature Computation and Statistical Metrics}\label{app:feature}
The ocular and cardiac features were calculated within fixed resampling windows, with corresponding formulas and statistical metrics summarised in Table~\ref{tab:eye_feature_metrics} and Table~\ref{tab:cv_feature_metrics}.

\begin{table*}
\centering
\caption{Description and Statistical Metrics of Eye-Tracking Features.}
\label{tab:eye_feature_metrics}
\newcolumntype{L}[1]{>{\raggedright\arraybackslash}p{#1}}
\begin{tabular}{@{}L{0.22\textwidth}L{0.5\textwidth}L{0.20\textwidth}@{}}
\toprule
\textbf{Eye Feature} & \textbf{Description with Formula} & \textbf{Statistical Metrics}\\
\midrule
Pupil Diameter &
Pupil diameter calculated separately for left, right, and both eyes within the resampling window. &
Mean, Min, Max, SD\\
\addlinespace

Saccade Amplitudes &
Angular distance of saccades in degrees. \newline
Formula: $\theta = 2 \arctan\left(\tfrac{d}{2D}\right)$, where $d$ is the pixel distance and $D$ is the viewing distance. &
Mean, Max\\
\addlinespace

Saccade Velocity &
Velocity of saccades in degrees/second. \newline
Formula: $v = \tfrac{\Delta \theta}{\Delta t}$. &
Mean, Max\\
\addlinespace

Saccade \& Fixation Count &
Event counts within each resampling window. &
Mean\\

Saccade \& Fixation Rate &
Per-second event rate. \newline
Formula: $\text{Rate} = \tfrac{\text{Event Count in Window}}{\text{Window Duration (s)}}$. &
Rate\\

Saccade/Fixation Ratio &
Ratio of total saccade duration to total fixation duration. \newline
Formula: $\text{SFR} = \tfrac{T_{\text{saccade}}}{T_{\text{fixation}}}$. &
Ratio\\
\addlinespace

Fixation Duration &
Average fixation duration. \newline
Formula: $FD = \tfrac{\sum_{i=1}^{N} t_i}{N}$, where $t_i$ is each fixation duration and $N$ the number of fixations. &
Mean, Max, Sum\\
\addlinespace

Area of Interest (AOI) &
AOIs defined from in-game screenshots by HSV colour filtering, edge detection, and contour extraction. Coordinates saved as pixel and normalised values. \newline
Portion Formula: $\text{AOI Portion} = \tfrac{\text{Samples in AOI}}{\text{Total Samples}}$. \newline
Rate Formula: $\text{AOI Rate} = \tfrac{\text{Entries into AOI}}{\text{Window Duration}}$. &
Portion, Rate\\
\bottomrule
\end{tabular}
\Description{}
\end{table*}

\begin{table*}
\centering
\caption{Description and Statistical Metrics of Cardiovascular Features.}
\label{tab:cv_feature_metrics}

\newcolumntype{L}[1]{>{\raggedright\arraybackslash\hspace{0pt}}p{#1}}

\begin{tabular}{@{}L{0.22\textwidth}L{0.5\textwidth}L{0.20\textwidth}@{}}
\toprule
\textbf{Cardiac Feature} & \textbf{Description with Formula} & \textbf{Statistical Metrics}\\
\midrule

Heart Rate (HR) & 
Number of heartbeats per window. Derived from interbeat intervals (IBIs). \newline
Formula: $\text{HR} = \frac{60}{\text{IBI}}$, where IBI is in seconds. &
Mean \\
\addlinespace

Heart Rate Variability (HRV) & 
Two indices computed from NN intervals. \newline
$\text{RMSSD} = \sqrt{\tfrac{1}{N-1} \sum_{i=1}^{N-1} (NN_{i+1} - NN_i)^2 }$ \newline
$\text{MeanNN} = \tfrac{1}{N} \sum_{i=1}^N NN_i$ &
Mean \\
\addlinespace

Blood Volume Pulse (BVP) Statistics & 
Signal statistics across a window. \newline
Mean: $\mu = \tfrac{1}{N}\sum_{i=1}^N x_i$ \newline
Std: $\sigma = \sqrt{\tfrac{1}{N}\sum_{i=1}^N (x_i-\mu)^2}$ &
Mean, Std \\
\bottomrule
\end{tabular}

\Description{}
\end{table*}

\section{Extracted Features and Final Feature Sets for Ocular and Cardiac Modalities}
\label{app:feature_summary}
Table~\ref{tab:feature_sets} summarises the complete set of ocular and cardiac features extracted from the raw signals, together with the final subset retained for model training. The complete set includes all features produced during preprocessing, whereas the final set contains only the measures retained after eliminating variables with low discriminative power or high collinearity.

\begin{table*}
\centering
\caption{Complete extracted feature set and final features used for each modality.}
\label{tab:feature_sets}
\begin{tabularx}{\textwidth}{l X X}
\toprule
\textbf{Modality} & \textbf{Complete Extracted Feature Set} & \textbf{Final Features Used} \\
\midrule

\textbf{Ocular} & 
\textbf{Pupil:} Left/Right/Avg diameter (Mean, Std, Min, Max); \newline
\textbf{Saccade:} Count mean; Rate; Amplitude (Mean, Max); Velocity max; Saccade to fixation ratio; \newline
\textbf{Fixation:} Duration (Max, Mean, Sum); \newline
\textbf{AOI:} Hand cards (Proportion, Rate); Top bar potion (Proportion, Rate)
&
\textbf{Saccade:} Count mean; Rate; Amplitude (Mean, Max); Velocity max; Saccade to fixation ratio; \newline
\textbf{AOI:} Hand cards (Proportion, Rate); Top bar potion (Proportion, Rate)
\\
\midrule

\textbf{Cardiac} &
\textbf{BVP Stats:} Min; Max; Mean; Std; Variance; RMS; Skewness; Kurtosis; \newline
\textbf{HRV:} Mean NN; RMSSD; pNN50; pNN20; Ratio
&
\textbf{Heart Rate:} Mean Heart Rate; Heart Rate Range
\\

\bottomrule
\end{tabularx}
\end{table*}

\section{Areas of Interest for Gaze Features}
\label{app:aoi}
\begin{figure*}
    \centering
    \includegraphics[width=0.9\textwidth]{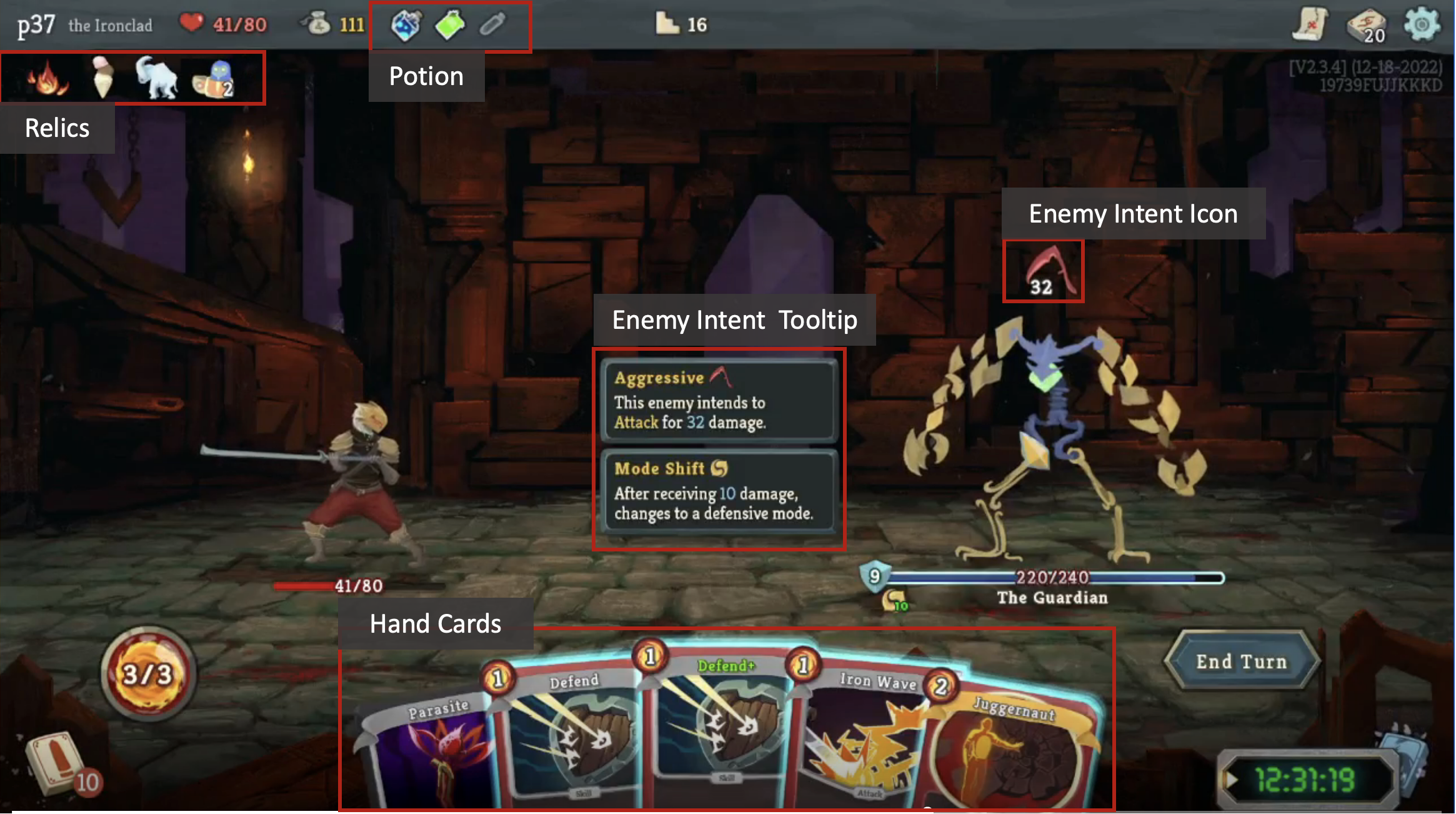}
    \caption{Areas of interest (AOIs) defined for gaze feature extraction in \textit{Slay the Spire}. AOIs included (1) hand cards at the bottom centre (decision options), (2) relics at the top-left (passive bonuses), (3) potions at the top (consumable items), and (4) enemy intent indicators above the opponent, including both the icon and tooltip (upcoming enemy action). These regions provided the basis for deriving gaze features such as fixation proportion and dwell time.}
    \label{fig:aoi}
    \Description[AOIs feature illustration]{The figure shows a combat screen from \textit{Slay the Spire} with AOIs outlined in red. At the bottom centre are hand cards, representing actions available to the player. At the top-left are relic icons, which provide passive effects. At the top are potion slots, showing consumable items. On the right, above the enemy, the intent icon and tooltip indicate the opponent’s next planned move. These AOIs were used to quantify gaze allocation to task-relevant interface elements.}
\end{figure*}

Table \ref{tab:interface_elements} documents the areas of interest (AOIs) used for gaze feature extraction in \textit{Slay the Spire}. Figure \ref{fig:aoi} illustrates the annotated interface, and the table summarises each AOI element with its function and screen position. These materials provide a transparent basis for how gaze allocation features, such as fixation proportion and dwell time, were derived.

\begin{table*}[t]
\centering
\caption{Key interface elements in \textit{Slay the Spire} with their positions and functions.}
\label{tab:interface_elements}

\newcolumntype{L}[1]{>{\raggedright\arraybackslash\hspace{0pt}}p{#1}}

\begin{tabular}{@{}L{0.16\textwidth}L{0.5\textwidth}L{0.30\textwidth}@{}}
\toprule
\textbf{Element} & \textbf{Function / Effect} & \textbf{Typical On-Screen Position} \\
\midrule
Relics & Permanent passive bonuses that affect the run (e.g., extra energy, healing, card draw). & Top-left area of the screen \\
Potions & One-time use consumable items used during combat to grant effects (e.g., heal, buff, damage). & Top-middle of the combat interface, in potion slots \\
Hand Cards & The cards the player currently can play, showing available actions (Attacks, Skills, Powers). & Bottom centre of the screen \\
Enemy Tooltip & Shows what the enemy intends to do next (e.g., attack amount, defense, buffs/debuffs), guiding player strategy. & Above the enemy on the combat screen \\
\bottomrule
\end{tabular}
\end{table*}

\section{Feature Importance Results}\label{app:predictor_score}
Table~\ref{tab:eye_catboost_importance} summarises the mean CatBoost feature importance across LOSO folds for the ocular modality, and Table~\ref{tab:xgb_cardiac_importance} summarises the mean XGBoost feature importance across LOSO folds for the cardiac modality.

\begin{table}
\centering
\caption{CatBoost feature importance for the ocular modality (mean across LOSO folds).}
\label{tab:eye_catboost_importance}
\begin{tabular}{l r}
\toprule
\textbf{Feature} & \textbf{Importance} \\
\midrule
Potion bar (AOI proportion) & 45.99 \\
Potion bar (AOI rate)       & 40.88 \\
Hand cards (AOI proportion) &  4.29 \\
Hand cards (AOI rate)       &  3.49 \\
Saccade--fixation ratio      &  2.36 \\
Saccade rate                &  1.51 \\
Mean saccade amplitude (deg) & 0.54 \\
Max saccade amplitude (deg)  & 0.40 \\
Max saccade velocity         & 0.36 \\
Mean saccade count           & 0.18 \\
\bottomrule
\end{tabular}
\end{table}

\begin{table}
\centering
\caption{XGBoost feature importance for the Cardiac modality (mean across LOSO folds).}
\label{tab:xgb_cardiac_importance}
\begin{tabular}{l c}
\toprule
\textbf{Feature} & \textbf{Importance} \\
\midrule
HR Range        & 0.56 \\
Mean Heart Rate  & 0.44 \\
\bottomrule
\end{tabular}
\end{table}

\clearpage

\end{document}